\documentclass[acmsmall,screen]{acmart}

\AtBeginDocument{\providecommand\BibTeX{{\normalfont B\kern-0.5em{\scshape i\kern-0.25em b}\kern-0.8em\TeX}}}

\setcopyright{acmcopyright}
\acmYear{2022}
\acmPrice{15.00}
\acmDOI{10.1145/3514260}

\acmJournal{TELO}
\acmVolume{1}
\acmNumber{1}
\acmArticle{1}
\acmMonth{1}

\acmSubmissionID{TELO-2021-15}

\usepackage{algorithm}
\usepackage{algpseudocode}
\algnewcommand{\IfThen}[2]{\State \algorithmicif\ #1\ \algorithmicthen\ #2}

\usepackage[utf8]{inputenc}
\usepackage{microtype}
\usepackage{siunitx}
\usepackage{textcomp}
\usepackage{textgreek}
\usepackage{times}

\usepackage{graphicx}
\usepackage[labelformat=simple]{subcaption}

\usepackage{amsmath}

\usepackage{amssymb}
\usepackage{bbm}
\usepackage{mathrsfs}
\usepackage{mathtools}
\usepackage{xfrac}

\usepackage{array}
\usepackage{booktabs}
\usepackage{multirow}
\usepackage{tabu}
\usepackage{tabularx}
\usepackage{tabulary}

\usepackage{color}
\usepackage{changepage} 

\newcommand{\atwoc}{\text{A2C}}
\newcommand{\curiosity}{\textit{Curiosity}}
\newcommand{\ppo}{\text{PPO}}

\newcommand{\asteroids}{\textit{Asteroids}}
\newcommand{\beamrider}{\textit{Beamrider}}
\newcommand{\breakout}{\textit{Breakout}}
\newcommand{\kungfumaster}{\textit{Kung-Fu~Master}}
\newcommand{\mspacman}{\textit{Ms.~Pac-Man}}
\newcommand{\pong}{\textit{Pong}}
\newcommand{\qbert}{\textit{Q*bert}}
\newcommand{\seaquest}{\textit{Seaquest}}
\newcommand{\atarienduro}{\textit{Enduro}}

\newcommand{\dense}{\textit{Dense}}
\newcommand{\sparse}{\textit{Sparse}}
\newcommand{\verysparse}{\textit{Very~Sparse}}
\newcommand{\supersparse}{\textit{Super~Sparse}}
\newcommand{\denseabbr}{\text{D}}

\newcommand{\atarifull}{\textit{Atari~2600}}
\newcommand{\vizdoom}{\textit{ViZDoom}}

\newcommand{\dismacro}[1]{\hat{m}_{#1}}

\newcommand{\methodrepeat}{\text{Repeat}}

\newcommand{\primspace}{\mathcal{A}}
\newcommand{\augspace}{\mathcal{M}}
\newcommand{\algo}{\mathcal{R}}
\newcommand{\env}{\mathcal{E}}
\newcommand{\statespace}{\mathcal{S}}
\newcommand{\policy}{\nu}

\newcommand{\fixtypocolor}{black}
\newcommand{\revisioncolor}{black}
\newcommand{\additioncolor}{black}

\newcommand{\fixtypo}[1]{\textcolor{\fixtypocolor}{#1}}
\newcommand{\revision}[1]{\textcolor{\revisioncolor}{#1}}

\begin{document}

\title{Reusability and Transferability of Macro Actions for Reinforcement Learning}

\author{Yi-Hsiang Chang}
\affiliation{\institution{National Tsing Hua University}
  \city{Hsinchu City}
  \country{Taiwan}}
\email{shawn420@gapp.nthu.edu.tw}

\author{Kuan-Yu Chang}
\affiliation{\institution{National Tsing Hua University}
  \city{Hsinchu City}
  \country{Taiwan}}
\email{s102020009@gapp.nthu.edu.tw}

\author{Henry Kuo}
\affiliation{\institution{Harvard University}
  \city{Cambridge}
  \country{USA}}
\email{hkuo@college.harvard.edu}

\author{Chun-Yi Lee}
\affiliation{\institution{National Tsing Hua University}
  \city{Hsinchu City}
  \country{Taiwan}}
\email{cylee@gapp.nthu.edu.tw}

\renewcommand{\shortauthors}{Y.-H. Chang, K.-Y. Chang, H. Kuo, and C.-Y. Lee}

\begin{abstract}
Conventional reinforcement learning (RL) typically determines an appropriate primitive action at each timestep. However, by using a proper macro action, defined as a sequence of primitive actions, an RL agent is able to bypass intermediate states to a farther state and facilitate its learning procedure. The problem we would like to investigate is what associated beneficial properties that macro actions may possess. In this paper, we unveil the properties of \textit{reusability} and \textit{transferability} of macro actions. The first property, \textit{reusability}, means that a macro action derived along with one RL method can be reused by another RL method for training, while the second one, \textit{transferability}, indicates that a macro action can be utilized for training agents in similar environments with different reward settings. In our experiments, we first derive macro actions along with RL methods. We then provide a set of analyses to reveal the properties of \textit{reusability} and \textit{transferability} of the derived macro actions.
\end{abstract}



\maketitle

\section{Introduction}
\label{sec:introduction}

Reinforcement learning (RL)~\cite{sutton1998introduction} has been shown to demonstrate super-human performance on a variety of environments and tasks~\cite{moriarty1999evolutionary,mnih2013playing,mnih2015human,mnih2016asynchronous,salimans2017evolution,schulman2017proximal,such2017deep}. In conventional methods, agents are restricted to make decisions at each timestep. However, the policies of RL agents intrinsically favor short-term goals due to reward discounting. Such situation is further exacerbated by the greedy nature of the agents which simply follow the policy and/or value functions. Therefore, researchers in the past years have proposed a few techniques to generate macro actions~\cite{durugkar2016deep,xu2019macro,Heecheol2019MacroAR}. A macro action (or simply ``\textit{a macro}'') is defined as an open-loop~\cite{distefano2012feedback} policy composed of a finite sequence of primitive actions. Once a macro is chosen, the actions will be taken by the agent without any further decision making process. Unfortunately, little attention has been paid to investigate the essential effects and the associated properties important for good macros in RL.

\revision{Macro actions work because of the \textit{embedding effect} and the \textit{evaluation effect} discussed in~\cite{botea2005macro}. The former enables bypassing a series of successor states from a start state that would normally be achieved in several steps, and thus allows the search space to be changed as well as the search depth to be reduced.  Since these bypassed states do not have to be evaluated, the search costs can also be reduced considerably.
The latter improves the search guidance by protecting the intermediate states from the greediness of the RL agents while executing them. Such a temporal abstraction can hardly be discovered using a single reward, or achieved by a single action, e.g., risking into a dangerous zone to retrieve a valuable item.} As macro actions \revision{that potentially possess these benefits} are allowed to have different lengths and arbitrary compositions of primitive actions, such diversified macro actions essentially form an enormous space.
We hereby define this space as the \textit{macro action space} (or simply ``\textit{macro space}'').   For a specific task in an environment, there exists good macros and bad macros in the macro space. Different macro actions have different impacts on an agent. A bad macro may lead the agent to undesirable states.  On the other hand, a good macro enables an RL agent to bypass multiple intermediate states and reach a target state quicker and more easily.  Such macro actions may require the agent to temporarily execute an action that may hurt the performance in the short term, while allowing it to achieve a long-term gain.

In addition, we further assume that good macros also exhibit invariance among different RL methods and similar environments.  A good macro should be able to be utilized by agents trained using different RL methods.  Furthermore, it should also be beneficial in similar environments with different reward settings. 
The first contribution of this research is a workflow that generates macros by a macro generation method.  We then show that two contemporary RL methods are benefited from the derived macros due to the above two effects.
The next contribution is that we evaluate the \textit{reusability} property of the derived macros between different RL methods.  This property allows the derived macros to be reused by different RL methods and benefit them. The third contribution is to reveal that the derived macro action possesses the property of \textit{transferability}. This property allows the derived macro actions to be utilized in similar task environments with different reward settings.  To the best of our knowledge, these two properties and the two effects of macros have not been properly discussed in the RL domain.

The paper is organized as follows. Section~\ref{sec:previous_works} briefly reviews the previous works. Sections~\ref{sec:problem_formulation}~and~\ref{sec:rl_background} describes the preliminaries. Section~\ref{sec:proposed_approach} explains our workflow. Section~\ref{sec:experimental_setup} details our experimental setups. Section~\ref{sec:compatibility} examines the \textit{embedding} and the \textit{evaluation effects}. Section~\ref{sec:transferability_reusability} discusses the \textit{reusability} and \textit{transferability} properties. Section~\ref{sec:conclusions} concludes this paper.
 \section{Previous Works}
\label{sec:previous_works}

The concept of macro actions has been adopted in the domain of planning~\cite{botea2005macro,sacerdoti1974planning,korf1985macro,dejong1986explanation,kaelbling1993hierarchical,newton2007learning,coles2007marvin,asai2015solving,chrpa2019improving,khetarpal2020options}, and has been shown to be able to provide advantages such as the \textit{embedding effect} and \textit{evaluation effect}~\cite{botea2005macro}.  The former enables bypassing a series of successor states from a start state, and thus allows the search space to be changed as well as the search depth to be reduced.  The latter allows the evaluation of a state to be different from the methods only based on primitive actions.  The two effects have been validated in the field of planning, however, have not yet been properly investigated in the domain of RL.  There have been few pioneering studies dedicated to developing macro actions for RL.  Previous researchers either produce macros in handcrafted manners~\cite{xu2019macro}, or derive them from expert demonstrations~\cite{Heecheol2019MacroAR}. The authors in~\cite{xu2019macro} showed that handcrafted macros can speed up their training processes in certain tasks. On the other hand, the authors in~\cite{Heecheol2019MacroAR} generated their macros from expert demonstrations via a variational auto-encoder.  Although these approaches have shown that macro actions are suitable for specific algorithms or applicable to grid-worlds, however, further properties such as \textit{reusability} and \textit{transferability} of macro actions have not been investigated and discussed.
 \section{Preliminaries}
\label{sec:problem_formulation}

In this section, we first provide the definition of macro actions.  Then, we provide a model of the environment permitting macro actions, which is a special case of Semi-Markov Decision Processes (SMDP)~\cite{sutton1999between}.  Next, we reformulate the essential equations in RL with macro actions, and provide the derivation procedure of them.

\paragraph{Macro action.}~A macro $m$ is defined as a finite sequence of primitive actions $m = (a_{1}, \ldots, a_{l})$, for all $a_i$ in an action space $\mathcal{A}$, and some natural number $l$.  The set of macros form a macro  space $\mathfrak{M} = \primspace{}^{+}$, where `+' stands for \textit{Kleene plus}.

\paragraph{Environment modeled.}~The environment we concern can be modeled as a special case of SMDP, which can be represented as a 4-tuple $(\statespace{}, \augspace{}, p_{ss'}^{m}, r_{s}^{m})$, where $\statespace{}$ is a set of states, $\augspace{}$ the finite set containing a single macro $m$ and all primitive actions contained in $\primspace{}$ provided by the environment $\env{}$, $p_{ss'}^{m}$ the transition probability from $s$ to $s'$ when executing $m$, and $r_{s}^{m}$ the reward received by the agent after executing $m$.  The expressions of $\augspace{}$, $p_{ss'}^{m}$, and $r_{s}^{m}$ are formulated as:
\begin{align}
    \augspace{} & = \primspace{} \cup \{ m \} \text{, where } m \in \mathfrak{M} \label{eq:augmented_action_space} \\
    p_{ss'}^{m} & = \mathbb{P} \left\{ s_{t+|m|} = s' \middle| s_{t} = s, m_{t} = m \right\} \label{eq:env_dynamic} \\
    r_{s}^{m} & = \mathbb{E} \left\{ \smashoperator{\sum_{\tau = 0}^{\ |m|-1}}{ r_{t+\tau}} \middle| s_{t} = s, m_{t} = m \right\} \label{eq:expected_reward} \\ \nonumber
\end{align}

\paragraph{Reinforcement learning.}~An RL agent interacts with $\env{}$ under a policy $\policy$ using provided $\augspace{}$, where $\policy$ is a mapping, $\policy: \statespace{} \times \augspace{} \to [0, 1]$.  The expected cumulative rewards it receives from each state $s$ under $\policy$ can be denoted as $V^{\policy}(s)$.  The objective of RL is to train an agent to learn an optimal policy such that it is able to maximize its expected return. The maximal expected return from each state $s$ under the optimal policy can be denoted as $V_{\augspace{}}^{*}(s)$. The expressions of $V^{\policy}$ and $V_{\augspace{}}^{*}$ can be represented as Eqs.~\eqref{eq:value_function}~and~\eqref{eq:optimal_value_function}, respectively, where the discount factor $\gamma \in [0, 1]$.  As a macro action is selected atomically as one of the actions by the agent, we modify the conventional formulation such that discounting is applied between macros rather than between the primitive actions within a macro. This encourages the agent to prefer to executing the provided macro action rather than consecutively performing a series of primitive actions when rewards are positive, while discouraging it when rewards are negative.
\begin{align}
    V^{\nu}(s) & = \sum_{m \in \augspace{}}{\nu(s, m) \left[ r_{s}^{m} + \gamma \smashoperator{\sum_{s' \in \statespace{}}}{p_{ss'}^{m}V^{\nu}(s')} \right]} \label{eq:value_function} \\
    V^{*}_{\augspace{}}(s) & = \max_{m \in \augspace{}}{ \left[ r_{s}^{m} + \gamma \sum_{s' \in \statespace{}}{p_{ss'}^{m} V^{*}_{\augspace{}}(s')} \right]} \label{eq:optimal_value_function}
\end{align}
 \revision{The expected cumulative rewards that an RL agent receives from each $s$ under $\policy{}$ after taking $m$ can be denoted as $Q^{\policy}(s, m)$. Furthermore, the maximal expected return from each state-macro pair $(s, m)$ under the optimal policy can be denoted as $Q_{\augspace{}}^{*}(s, m)$. The expressions of $Q^{\policy}$ and $Q_{\augspace{}}^{*}$ can be represented as Eqs.~\eqref{eq:q_value_function}~and~\eqref{eq:q_optimal_value_function}, respectively.}

\begin{align}
    Q^{\nu}(s, m) & = r_{s}^{m} + \gamma \sum_{s' \in \mathcal{S}}{p_{ss'}^{m} \sum_{m' \in \mathcal{M}}{\nu(s', m') Q^{\nu}(s', m')}} \label{eq:q_value_function} \\
    Q_{\mathcal{M}}^{*}(s, m) & = r_{s}^{m} + \sum_{s' \in \mathcal{S}}{p_{ss'}^{m} \max_{m' \in \mathcal{M}}{Q_{\mathcal{M}}^{*}(s', m')}} \label{eq:q_optimal_value_function}
\end{align}
 \color{\additioncolor}
\section{Background Materials of Deep Reinforcement Learning}
\label{sec:rl_background}

In this section, we introduce the background materials of Deep RL (DRL) methods adopted in this work. The goal of a DRL agent is to maximize its accumulated long-term rewards $G_{t}$ over discrete timesteps. Unlike an RL agent, the policy $\pi_{\theta}(a|s)$ of a DRL agent is a conditional probability distribution parameterized by a deep neural network (DNN) with parameters $\theta$. 
Since training agents using off-policy based methods (e.g., Deep Q-Network~(DQN)~\cite{mnih2015human}) is much slower than training agents based on on-policy methods~\cite{stooke2018accelerated}, the DRL methods considered in this work include two on-policy based methods Advantage Actor-Critic (A2C)~\cite{mnih2016asynchronous} and Proximal Policy Optimization (PPO)~\cite{schulman2017proximal}, as well as the curiosity-driven exploration technique~\cite{pathak2017curiosity}.  As A2C and PPO also train their Q-functions (i.e., like DQN) to serve as the critics, it is more feasible to leverage the advantage offered by them to quickly validate the properties of macros for multiple environments. These DRL methods (i.e., $\atwoc{}$ and $\ppo{}$) have been widely adopted as the baseline approaches in the literature and are capable of completing the same training timesteps with less amount of physical time~\cite{stooke2018accelerated}. Among these methods, A2C and PPO utilize the policy gradient methods~\citep{sutton2000policy,williams1992simple} to train the agents, which directly optimize $\theta$ in the direction of $\nabla_\theta \mathbb{E}_{a_t \sim \pi_{\theta}} \left[\log \pi_{\theta}(a_t|s_t) \, G_t \right]$ or $\nabla_\theta \mathbb{E}_{a_t \sim \pi_{\theta}} \left[\log \pi_{\theta}(a_t|s_t) \, A \right]$, where the latter is adopted for Actor-Critic based approaches and $A$ denotes the advantage function.

\subsection{Advantage Actor-Critic (A2C)}
\label{subsec:drl_atwoc}

Advantage Actor-Critic $\atwoc{}$ is a famous on-policy method, which updates the value function based on the samples generated by the current policy. It is a synchronous variant of Asynchronous Advantage Actor-Critic (A3C)~\cite{mnih2016asynchronous}, which trains agents in parallel on multiple instances of the environment. A2C offers better utilization of GPUs than A3C. 
For both A2C and A3C, a learning agent uses the predictions of the value function $V(s)$ (i.e., the \textit{Critic}) to update the policy function $\pi_{\theta}(a|s)$ (i.e., the \textit{Actor}). During the training phase, the critic is trained to estimate the value function $V(s)$, while the actor optimizes $\pi_{\theta}(a|s)$ by minimizing the loss term:
\begin{equation}
    L_{actor} = -\mathbb{E}_{s, a \sim \pi_{\theta}}\big[G_t - V(s) + \beta H(\pi_{\theta}(\cdot|s)) \big],
    \label{eq::a2closs}
\end{equation}
where $\beta$ is a parameter for controlling the strength of the entropy $H(\pi_{\theta}(\cdot|s))$. A2C utilizes $H(\pi_{\theta}(\cdot|s))$ to encourage exploration, and prevent agents from converging prematurely to a sub-optimal policy.

\subsection{Proximal Policy Optimization (PPO)}
\label{subsec:drl_ppo}

Proximal Policy Optimization ($\ppo{}$) is another famous on-policy method, which computes an update at every timestep that minimizes the cost function while ensuring the deviation from the previous policy is relatively small. One of the two main variants of PPO is based on a clipped surrogate objective expressed as:
\begin{equation}
L^{CLIP}(\theta) = \mathbb{E} \left[ \frac{\pi_{\theta}(a|s)}{\pi_{\theta_{old}}(a|s)}\hat{A}, \text{clip}\left(\frac{\pi_{\theta}(a|s)}{\pi_{\theta_{old}}(a|s)}, 1-\epsilon, 1+\epsilon)\hat{A}\right) \right],
\end{equation}
where $\hat{A}$ is the advantage estimate, and $\epsilon$ a hyperparameter. The clipped probability ratio is used to prevent large changes to the policy between updates. The other variant of PPO employs a different surrogate objective during the training phase.  This surrogate objective is based on an adaptive penalty derived from the KL divergence between an old policy $\pi_{\theta_{old}}$ and the current one $\pi_{\theta}(\cdot|s)$, given by:
\begin{equation}
L^{KLPEN}(\theta) = \mathbb{E} \left[ \frac{\pi_{\theta}(a|s)}{\pi_{\theta_{old}}(a|s)}\hat{A} - \beta KL\left[\pi_{\theta_{old}}(\cdot|s), \pi_{\theta}(\cdot|s)\right]\right],
\end{equation}
where $\beta$ is an adaptive coefficient adjusted according to the observed change in the KL divergence. In this work, we employ the former surrogate objective due to its better empirical performance.

\subsection{Curiosity-driven Exploration}
\label{subsec:curiosity}

Curiosity-driven exploration is an exploration strategy adopted by a number of DRL researchers in recent years~\cite{houthooft2016vime,bellemare2016unifying,ostrovski2017count,pathak2017curiosity,burda2018large,burda2018exploration} in order to explore environments more efficiently. While conventional random exploration strategies are easily trapped in local minima of state spaces for complex or sparse reward environments, curiosity-based methodologies tend to discover relatively un-visited regions, and therefore are likely to explore more effectively. In addition to the extrinsic rewards provided by the environments, most curiosity-driven exploration strategies introduce intrinsic rewards generated by the agent to encourage itself to explore novel states. The intrinsic rewards are formulated as the prediction errors of a forward dynamics network. Given an observation $o_t$, a feature representation $\phi_t$ is generated by an embedding network. The forward dynamics network $f$ is then used to predict the representation of the next observation $\phi_{t+1}$ by $\phi_t$ and the agent's taken action $a_t$. The mean-squared error (MSE) $||f(\phi_t, {a_t}) - \phi_{t+1})||^2$ is used as the intrinsic reward signal during the training phase, and serves as the training objective for $f$ to minimize. When the same observations are visited multiple times, their MSE errors drop, indicating that the agent are becoming familiar with them.  In this work, one of our baseline~\textit{Curiosity} is implemented based on the concept of curiosity-driven exploration.
 \color{black}
    \begin{figure}[t]
\centering
\includegraphics[width=.9\linewidth]{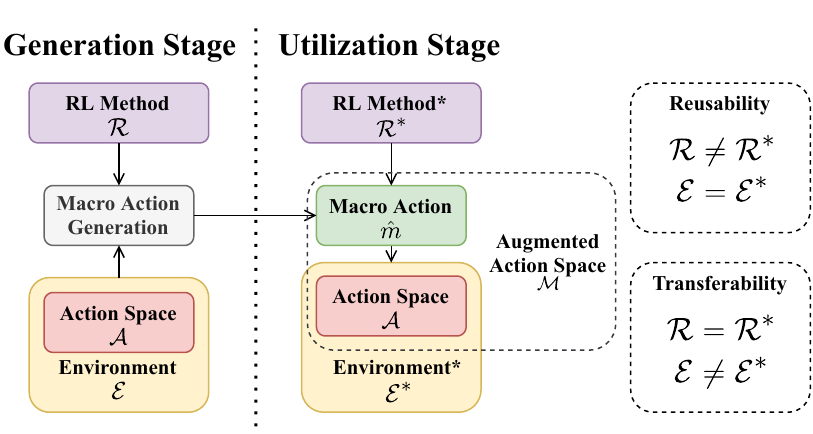}
\caption{
A illustration of the workflow adopted in this research.  It consists of a macro action generation stage and a macro action utilization stage.  The former aims to derive sufficiently good macros, while the latter utilizes the derived macros in various evaluation configurations.  The \textit{reusability} and \textit{transferability} properties concerned in this work are examined based on this workflow.
}
\label{fig:genetic_algorithm}
\end{figure}
     \begin{algorithm}[t]
\caption{The macro action generation stage}
\label{alg:genetic_algorithm}
\small
\begin{algorithmic}[1]
\State \textbf{input:} Environment $\env{}$
\State \textbf{input:} Reinforcement Learning algorithm $\algo{}$
\State \textbf{input:} Total number of constructed macros $k$
\State \textbf{input:} Number of macros retained after selection phase $q$
\State \textbf{input:} Number of macros mutated by append operator $q_{+}$
\State \textbf{input:} Number of macros mutated by alternation opertaor $q_{*}$
\State \textbf{output:} Best-performing macro $\hat{m}$
\Function{Construction}{$\env{}$, $\algo{}$, $k$, $q$, $q_{+}$, $q_{*}$}
    \State \textbf{initialize:} $g = q_{+} + q_{*}$
    \State \textbf{initialize:} $G = [m_1, \ldots, m_{g}]$, list of random macros \label{alg:genetic_algorithm:initialize}
    \State \textbf{initialize:} $F_{G} = [f_{1}, \ldots, f_{g}]$, list of fitness of $m$ in $G$
    \State \textbf{initialize:} $Q = []$, list of retained macros
    \State \textbf{initialize:} $F_{Q} = []$, list of fitness of $m$ in $Q$
    \State \textbf{initialize:} $\primspace{}$ = the primitive action space of $\env{}$
    \State \textbf{initialize:} $i=0$, number of constructed macro
    \While{$i<k$}
        \For{$j$ \textbf{in} $1$ \textbf{to} $g$} \label{alg:genetic_algorithm:fitness_start}
            \State $f_{j}=$ \Call{Fitness}{$\env{}$, $\algo{}$, $m_j$} \label{alg:genetic_algorithm:evaluate}
                \Comment{Algorithm~\ref{alg:evaluation}}
            \State $i=i+1$
            \IfThen{$i>=k$}{\textbf{break}}
        \EndFor \label{alg:genetic_algorithm:fitness_end}
        \State $Q$ = $[$top $q$ macros in $Q \cup G$ based on $F_{Q}$ and $F_{G}]$ \label{alg:genetic_algorithm:select_start}
        \State Update $F_{Q}$ \label{alg:genetic_algorithm:select_end}
        \IfThen{$i>=k$}{\textbf{break}}
        \State $Q_+=\emptyset,Q_{*} = \emptyset$
        \For{$m$ \textbf{in} $[q_+$ randomly selected macros from $Q]$} \label{alg:genetic_algorithm:mutate_start}
            \State $Q_{+}$ = $Q_{+} \cup \{\Call{Append}{\primspace{},m}\}$ \label{alg:genetic_algorithm:append} 
                \Comment{Algorithm~\ref{alg:genetic_append}}
        \EndFor
        \For{$m$ \textbf{in} $[q_*$ randomly selected macros from $Q]$}
            \State $Q_{*}$ = $Q_{*} \cup \{\Call{Alter}{\primspace{},m}\}$ \label{alg:genetic_algorithm:alteration} 
                \Comment{Algorithm~\ref{alg:genetic_alteration}}
        \EndFor
        \State $G=Q_+\cup Q_*$ \label{alg:genetic_algorithm:mutate_end}
    \EndWhile
    \State \textbf{return} The best-performing macro $\hat{m}$  in $Q$ based on $F_{Q}$
\EndFunction
\end{algorithmic}
\end{algorithm}
     \begin{algorithm}[t]
\caption{Fitness function}
\label{alg:evaluation}
\small
\begin{algorithmic}[1]
\State \textbf{input:} Environment $\env{}$
\State \textbf{input:} Reinforcement learning method $\algo{}$
\State \textbf{input:} Macro action $m$
\State \textbf{output:} Fitness score of $m$
\Function{Fitness}{$\env{}$, $\algo{}$, $m{}$}
    \State \textbf{initialize:} $\primspace{}$ = the primitive action space of $\env{}$
    \State \textbf{initialize:} $\augspace{}$ = $\primspace{} \cup \{ m \}$
    \State Learn a policy over $\augspace{}$ in $\env{}$ using $\algo{}$
    \State \textbf{return} Average of ``last 100 episode rewards''
\EndFunction
\end{algorithmic}
\end{algorithm}
     \begin{algorithm}[t]
\caption{Append operator}
\label{alg:genetic_append}
\small
\begin{algorithmic}[1]
\small
\State \textbf{input:} Primitive action space $\primspace{}$
\State \textbf{input:} Macro $m = (a_1,\ldots,a_{|m|})$
\State \textbf{output:} Mutated macro $m_{+}$
\Function{Append}{$\primspace{}$, $m$}
    \State \textbf{initialize:} $\alpha$ = random element in $\primspace{}$
    \State \textbf{return} $m_{+} = (a_1,\ldots,a_{|m|},\alpha)$
\EndFunction
\end{algorithmic}
\end{algorithm}
     \begin{algorithm}[t]
\caption{Alteration operator}
\label{alg:genetic_alteration}
\small
\begin{algorithmic}[1]
\small
\State \textbf{input:} Primitive action space $\primspace{}$
\State \textbf{input:} Macro $m = (a_1,\ldots,a_{|m|})$
\State \textbf{output:} Mutated macro $m_{*}$
\Function{Alter}{$\primspace{}$, $m$}
    \State \textbf{initialize:} $\alpha$ = random element in $\primspace{}$
    \State \textbf{return} $m_{*} = (\alpha,a_2,\ldots,a_{|m|})$
\EndFunction
\end{algorithmic}
\end{algorithm}
 \section{The Proposed Workflow}
\label{sec:proposed_approach}

In this section, we present the proposed workflow for investigating the \textit{reusability} and \textit{transferability} properties for macro actions, which is the main objective of this research.  Fig.~\ref{fig:genetic_algorithm} illustrates the workflow adopted in this paper. It contains two stages: a macro action generation stage, and a macro action utilization stage for performing various evaluations.  The former stage takes into account the RL method $\algo{}$, $\primspace{}$, and $\env{}$, and derives a macro action $\hat{m}$ that is sufficiently good enough for an RL agent to use as a means of performing temporal abstraction in $\env{}$. The latter stage then encapsulates the derived macro $\hat{m}$ and $\primspace{}$ in $\augspace{}$, and utilizes $\augspace{}$ in our evaluation experiments.  The workflow presented in Fig.~\ref{fig:genetic_algorithm} allows different experimental configurations to be defined.  

We formulate our macro action generation stage as Algorithm~\ref{alg:genetic_algorithm}, where the macro action construction method is based on a \textit{genetic algorithm} (GA)~\cite{mitchell1998introduction}. The promise of GA is that due to its simplicity, it offers an effective and easy way to construct sufficiently good macro actions.  As opposed to works that concentrate on evolutionary methods, our aim focuses on generating macro actions for investigating their properties, and is not proposing novel and general macro generation methods.  GA offers three promising advantages for our workflow. First, it eliminates the dependency of the macro action derivation procedure from human supervision. Second, it produces diversified macros by mutation. Third, it retains good macro actions from mutation over generations.

Algorithm~\ref{alg:genetic_algorithm} is established atop three modules: (1) the fitness function, (2) the append operator, and (3) the alteration operator. These three modules serve as essential roles in Algorithm~\ref{alg:genetic_algorithm}, and are additionally formulated as Algorithms~\ref{alg:evaluation},~\ref{alg:genetic_append}, and~\ref{alg:genetic_alteration}, respectively. Algorithm~\ref{alg:evaluation} evaluates a macro action $m$ by first appending it to $\primspace{}$ of the target $\env{}$ to form an augmented action space $\augspace{}$, and then measuring the fitness score of $m$ for an agent with $\augspace{}$ in $\env{}$ using a given RL method $\algo{}$ after training for a period. A single macro $m$ is evaluated rather than a set of macros, because a set of macros may cause ambiguity for determining the relative fitness of each macro in the set.

Our macro action generation stage consists of four phases: the ``\textit{initialization phase}'', the ``\textit{fitness phase}'', the ``\textit{selection phase}'', and the ``\textit{mutation phase}''.
We walk through Algorithm~\ref{alg:genetic_algorithm} and highlight the four phases of GA as the following. Line~\ref{alg:genetic_algorithm:initialize} corresponds to the ``\textit{initialization phase}'', which initializes the population of the macro actions with a number of randomly generated macro actions containing two primitive actions. Lines~\ref{alg:genetic_algorithm:fitness_start}-\ref{alg:genetic_algorithm:fitness_end} correspond to the ``\textit{fitness phase}'', \fixtypo{which performs fitness evaluation by Algorithm~\ref{alg:evaluation}.} Lines~\ref{alg:genetic_algorithm:select_start}-\ref{alg:genetic_algorithm:select_end} correspond to the ``\textit{selection phase}'', which retains the top performers in the population while eliminating the remaining ones from it. Lastly, lines~\ref{alg:genetic_algorithm:mutate_start}-\ref{alg:genetic_algorithm:mutate_end} correspond to the ``\textit{mutation phase}'', which randomly selects macro actions from the population, mutates them by Algorithms~\ref{alg:genetic_append}~and~\ref{alg:genetic_alteration}, and then form a new generation.

After the macro generation stage, a macro action $\hat{m}$ is derived. The macro utilization stage then uses RL method $\algo{}^{*}$, $\primspace{}$, and $\env{}^{*}$ to perform evaluations of $\hat{m}$. Three different configurations are considered in this research.  When the \textit{embedding effect} and the \textit{evaluation effect} are aimed to be evaluated, $\algo{}^{*}$ and $\env{}^{*}$ are set to be the same as $\algo{}$ and $\env{}$, respectively, such that these effects can be reflected in the evaluation experiments.  When evaluating the \textit{reusability} property, $\env{}^{*}$ is maintained to be the same as $\env{}$, while the RL method is changed to a different one. i.e., $\algo{} \neq \algo{}^{*}$ and $\env{} = \env{}^{*}$. Finally, when \textit{transferability} is to be examined, $\env{}^{*}$ is configured to be similar to $\env{}$ with a different reward setting under the same RL method, i.e., $\algo{}=\algo{}^{*}$ and $\env{} \neq \env{}^{*}$, as illustrated in Fig.~\ref{fig:genetic_algorithm}.     \begin{figure}[t]
\centering
\begin{subfigure}[t]{0.39\linewidth}
    \includegraphics[width=.9\linewidth]{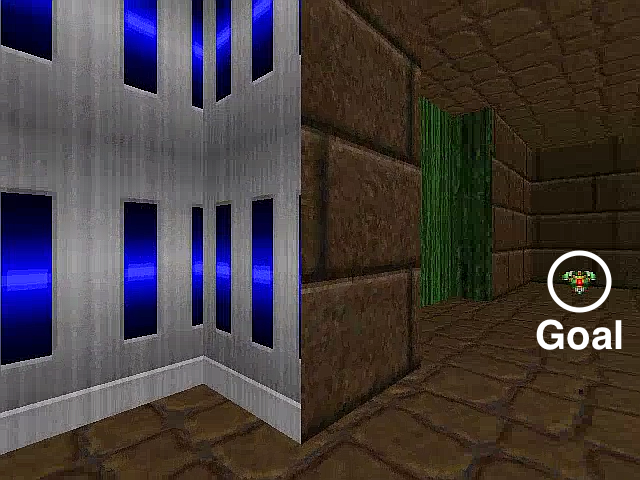}
    \subcaption{\vizdoom{}.}
    \label{fig:doom:snapshot}
\end{subfigure}
\begin{subfigure}[t]{0.6\linewidth}
    \includegraphics[width=.47\linewidth]{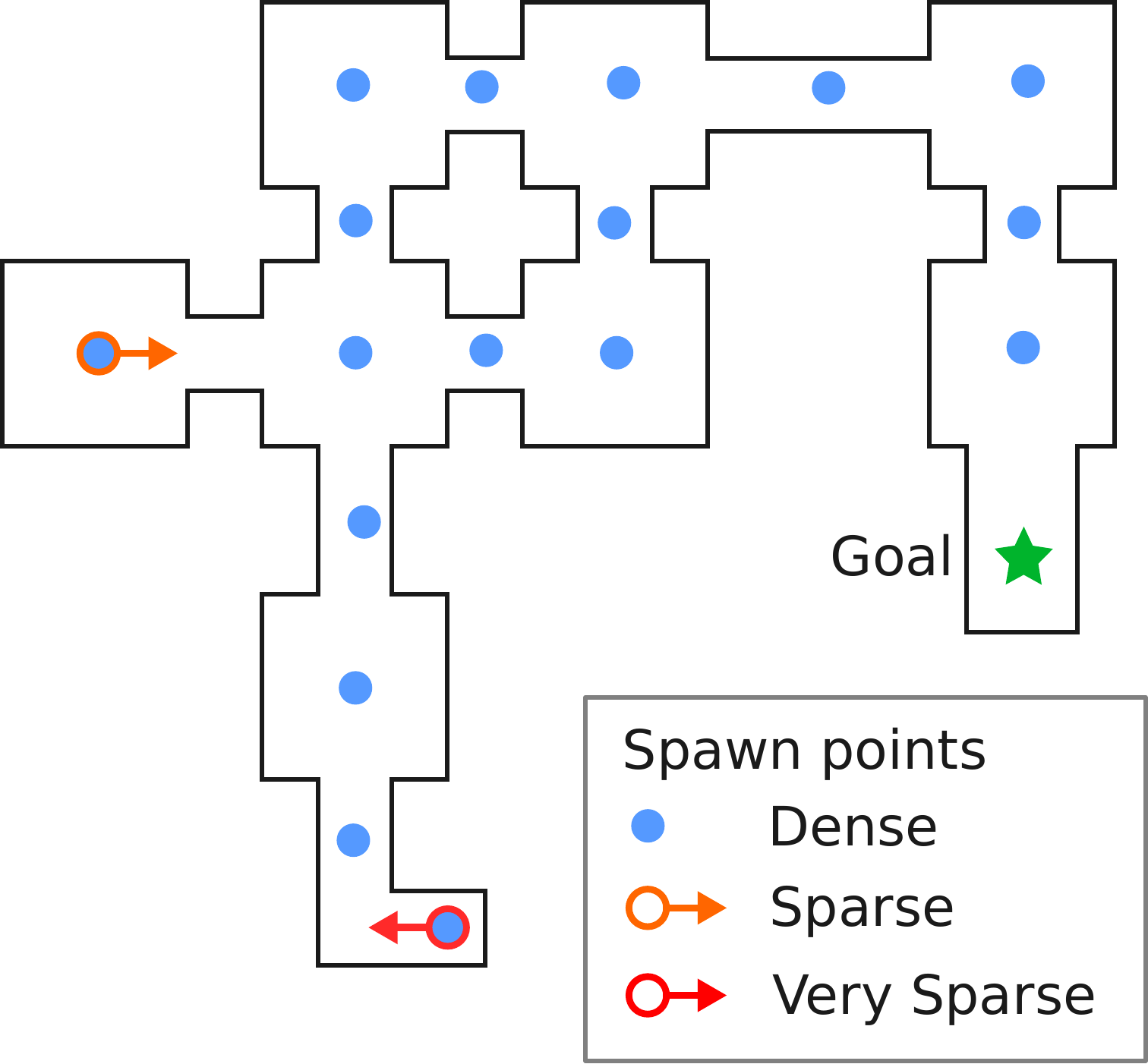}
    \includegraphics[width=.4\linewidth]{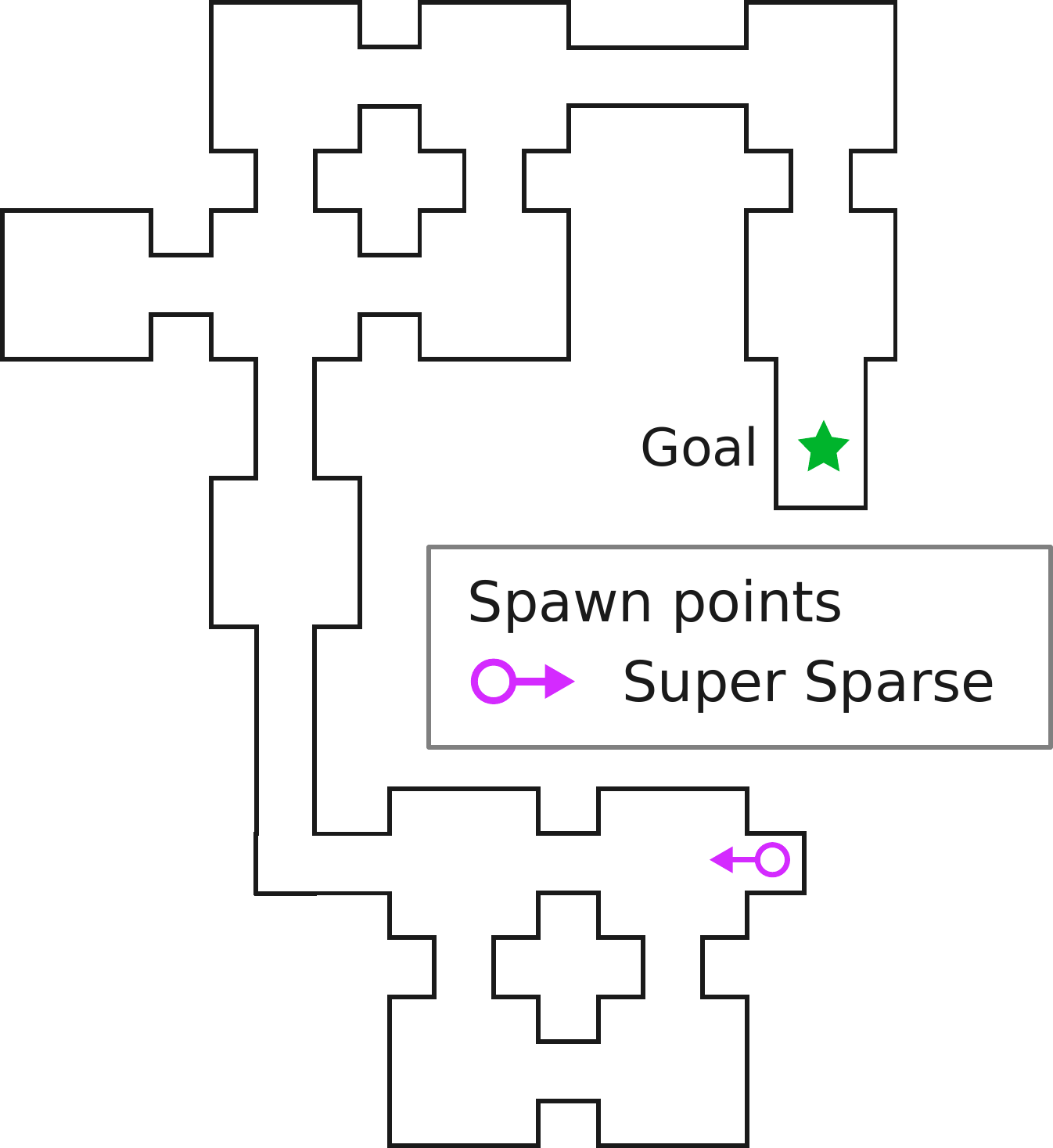}
    \subcaption{Map layouts of the \vizdoom{} tasks.}
    \label{fig:doom:map}
\end{subfigure}
\caption{\subref{fig:doom:snapshot} \vizdoom{} is a three-dimensional first-person perspective environment. The agents in \vizdoom{} make decisions solely based on their visual observations which do not contain any information about the locations of the agent and the goal. When performing rotations, an agent does not make direct 90-degree turns as in grid worlds. The turning angles are less than 90 degrees and hence, it is required to perform a sequence of turning actions before making a sharp turn. \subref{fig:doom:map} The blue points denote the random spawn points for the \dense{} task, while the orange, red, and purple arrows denote the fixed spawn points of the agents and their initial orientations for the \sparse{}, \verysparse{}, and \supersparse{} tasks, respectively.
}
\label{fig:doom}
\end{figure}
     \begin{table}[t]
\centering
\caption{List of the hyperparameters for the RL methods.}
\label{tab:hyperparameter}
\begin{tabular}{l|ccc}
\toprule
Hyperparameter                          & \atwoc{}       & \ppo{}       & \curiosity{} \\
\midrule
Discount factor                         & 0.99      & 0.99      & 0.99      \\
Number of frame skip                    & 4         & 4         & 4         \\
Number of parallel envs.                & 16        & 16        & 20        \\
Rollout length                          & 5         & 128       & 20        \\
Batch size                              & 80        & 2048      & 400       \\
Value function coefficient              & 0.25      & 0.5       & 0.5       \\
Entropy coefficient                     & 0.01      & 0.01      & 0.01      \\
Gradient clipping maximum               & 4.0       & 0.5       & 40.0      \\
Learning rate                           & 0.001     & 0.0003    & 0.0003    \\
Learning rate annealing                 & Linear    &           & Constant  \\
Clipping parameter                      &           & 0.2       &           \\

Policy gradient loss weight $\lambda$   &           &           & 0.1       \\
Forward model loss $\beta$              &           &           & 0.2       \\
Intrinsic reward scaling factor         &           &           & 0.01      \\
\bottomrule
\end{tabular}
\end{table}
 \section{Experimental Setups}
\label{sec:experimental_setup}

In this study, we used our customized machines to perform our experiments. In total, we utilized up to 64 CPU cores and six graphics cards in our experiments. In the following subsections, we first introduce the general setups. Then, we describe the environments and RL methods for our \textit{reusability} experiments. Finally, we explain the environments and RL method for our \textit{transferability} experiments.

\subsection{General Setup}
\label{subsec:general_setup}

All of the macro actions presented in this paper are derived by Algorithm~\ref{alg:genetic_algorithm}, if not specifically mentioned.  The four parameters used in Algorithm~\ref{alg:genetic_algorithm}, $k$, $q$, $q_{+}$, and $q_{*}$, are set to 50, 8, 5, and 3 throughout our experiments, respectively. The former two parameters are designed to stabilize the average fitness of the population,  while the latter two are selected such that GA prefers to append over alteration so as to encourage the growth of the length of our derived macro actions. The training time of ``fitness phase'' for Algorithm~\ref{alg:evaluation} is set to 5M, which is sufficient to determine whether the constructed macro action is satisfactory or not.  Please note that all of the curves presented in this paper are generated based on five different random seeds, and are drawn with 95\% confidence interval (displayed as the shaded areas).

\subsection{Setup for the Reusability Experiments}
\label{subsec:compatibility_transferability_setup}

\paragraph{Environments.}~We evaluate the derived macro actions on the following eight \atarifull{} \cite{bellemare13arcade} environments: \asteroids{}, \beamrider{}, \breakout{}, \kungfumaster{}, \mspacman, \pong{}, \qbert{}, and \seaquest{}. We first validate that our derived macro actions reflect the advantages of the \textit{embedding} and the \textit{evaluation effects} that benefit the selected RL methods in Section~\ref{sec:compatibility}. Then we present results and analyses for our \textit{reusability} experiments in Section~\ref{sec:transferability_reusability}.

\paragraph{RL methods.}~In our experiments, we select advantage actor-critic (\atwoc{})~\cite{mnih2016asynchronous} and proximal policy optimization (\ppo{})~\cite{schulman2017proximal} implementations from \cite{stable-baselines} as our RL methods for training the agents. The detailed hyperparameter setups are summarized in Table~\ref{tab:hyperparameter}.

\subsection{Setup for the Transferability Experiments}
\label{subsec:reusability_setup}

\paragraph{Environments.}~We employ \vizdoom{} as our environments for examining the \textit{transferability} property.  \vizdoom{} is a research platform featuring complex three-dimensional first-person perspective environments, as shown in Fig.~\ref{fig:doom:snapshot}. The agents in \vizdoom{} make decisions based on visual observations which do not contain any information about the locations of the agent and the goal. For \vizdoom{}, we evaluate our derived macro on the default task \texttt{my\_way\_home} (denoted as ``\dense{}''). Then we demonstrate that the macro benefits the selected RL method in Section~\ref{sec:compatibility}.  We further use the ``\sparse{}'', ``\verysparse{}'', and ``\supersparse{}'' (developed by us) tasks for analyzing the \textit{transferability} property of the constructed macro in Section~\ref{sec:transferability_reusability}.  The \supersparse{} task comes with extended rooms and corridors in which the distance between the spawn point of the agent and the goal is farther than the other tasks.  The map layouts for these tasks are depicted in Fig.~\ref{fig:doom}. When performing rotations, an agent does not make direct 90-degree turns as in grid worlds. The turning angles are less than 90 degrees and hence, it is required to perform a sequence of turning actions before making a sharp turn.

\paragraph{RL method.}~We implemented an intrinsic curiosity module (ICM)~\cite{pathak2017curiosity} along with \atwoc{} (together denoted as ``\curiosity{}''). The hyperparameters for ``\curiosity{}'' are summarized in Table~\ref{tab:hyperparameter}.
     \begin{figure}[t]
\centering
\begin{subfigure}[b]{.33\linewidth}
    \includegraphics[width=.9\linewidth]{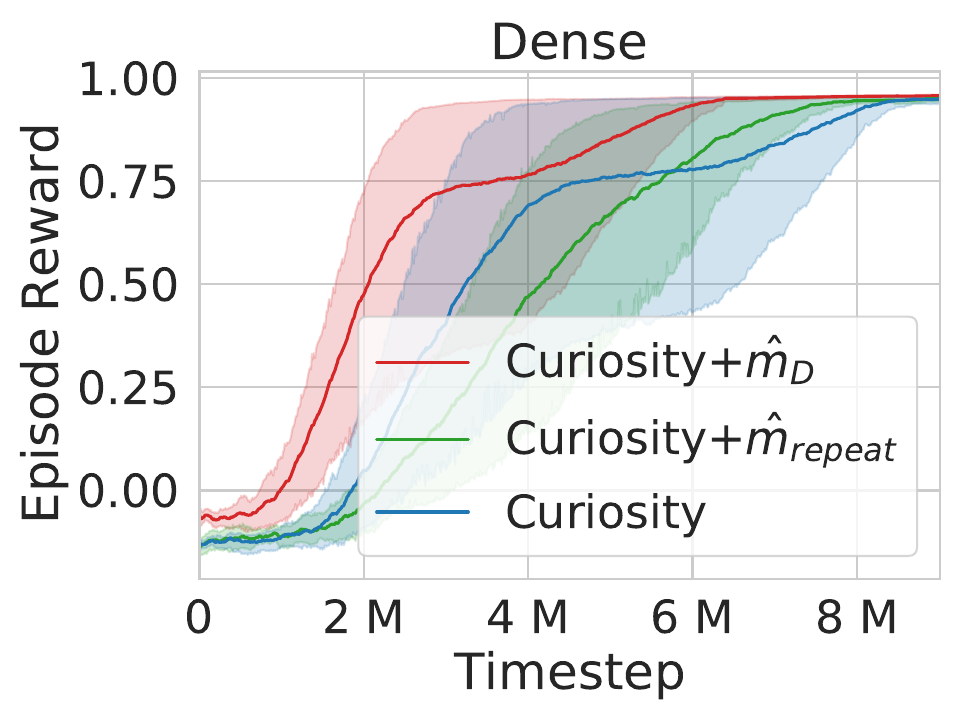}
    \subcaption{\dense{}.}
    \label{fig:learning_curves:macro_example:dense}
\end{subfigure}
\begin{subfigure}[b]{.33\linewidth}
    \includegraphics[width=.9\linewidth]{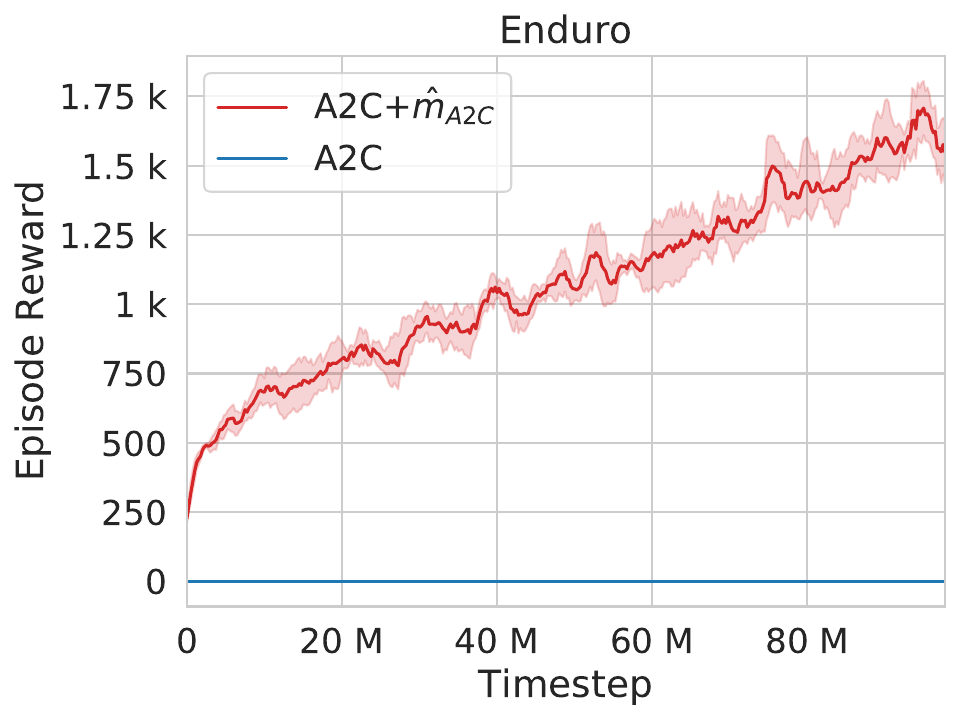}
    \subcaption{\atarienduro{}.}
    \label{fig:learning_curves:macro_example:enduro}
\end{subfigure}
\caption{The learning curves w/ and w/o the derived macros.}
\label{fig:learning_curves:macro_example}
\end{figure}
     \begin{figure*}[t]
\centering
\begin{subfigure}[b]{0.24\linewidth}
    \includegraphics[width=.9\linewidth]{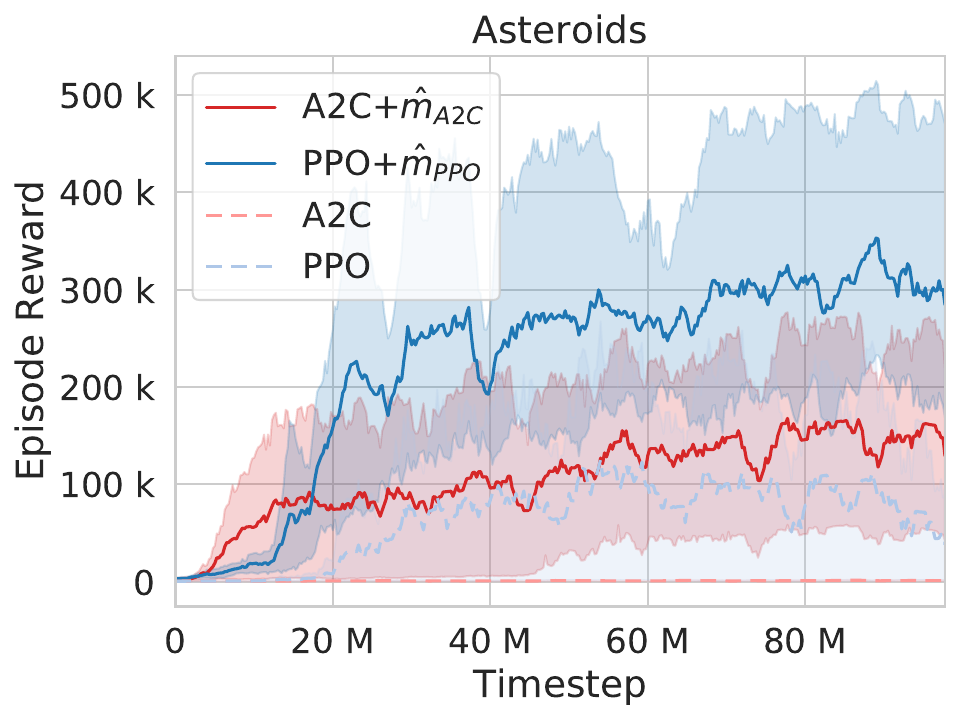}
    \label{fig:learning_curves:compatibility:asteroids}
\end{subfigure}
\begin{subfigure}[b]{0.24\linewidth}
    \includegraphics[width=.9\linewidth]{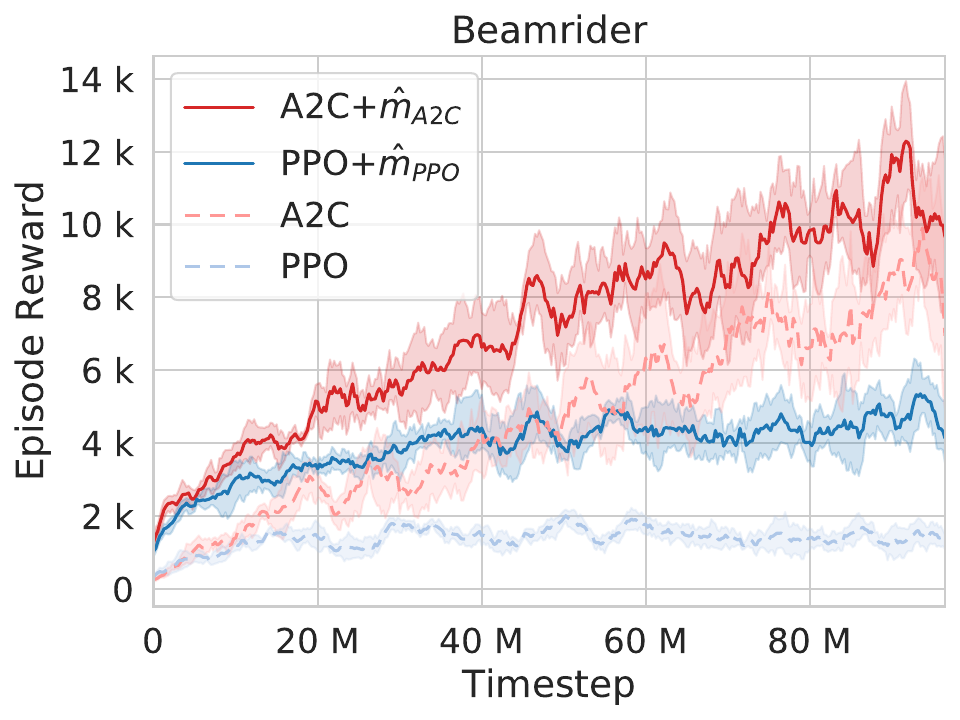}
    \label{fig:learning_curves:beam_rider}
\end{subfigure}
\begin{subfigure}[b]{0.24\linewidth}
    \includegraphics[width=.9\linewidth]{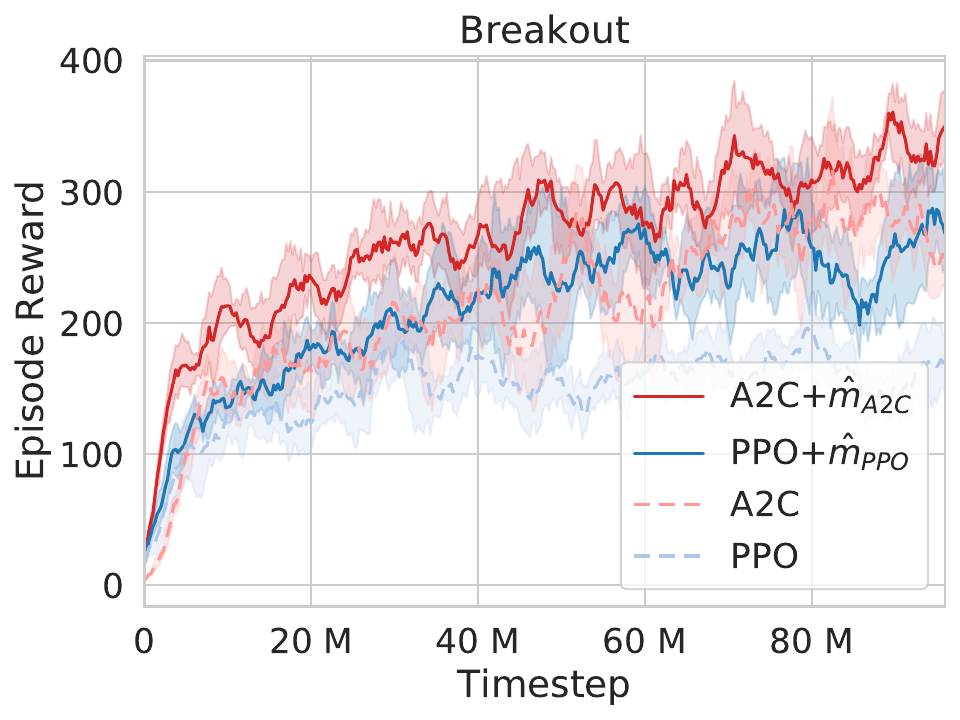}
    \label{fig:learning_curves:compatibility:breakout}
\end{subfigure}
\begin{subfigure}[b]{0.24\linewidth}
    \includegraphics[width=.9\linewidth]{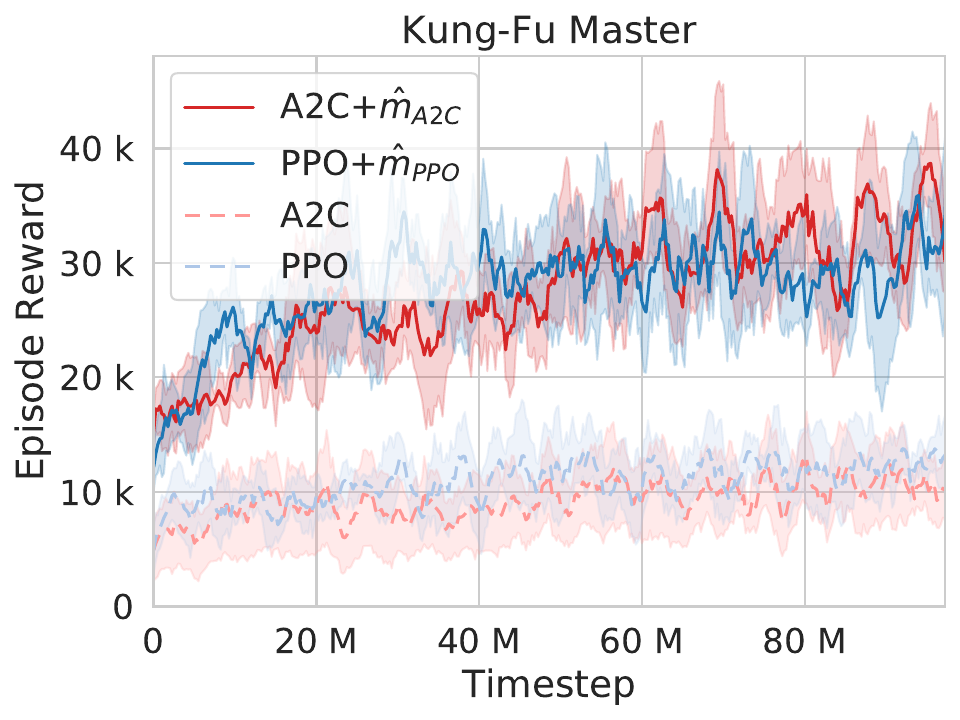}
    \label{fig:learning_curves:compatibility:kung_fu_master}
\end{subfigure}
\begin{subfigure}[b]{0.24\linewidth}
    \includegraphics[width=.9\linewidth]{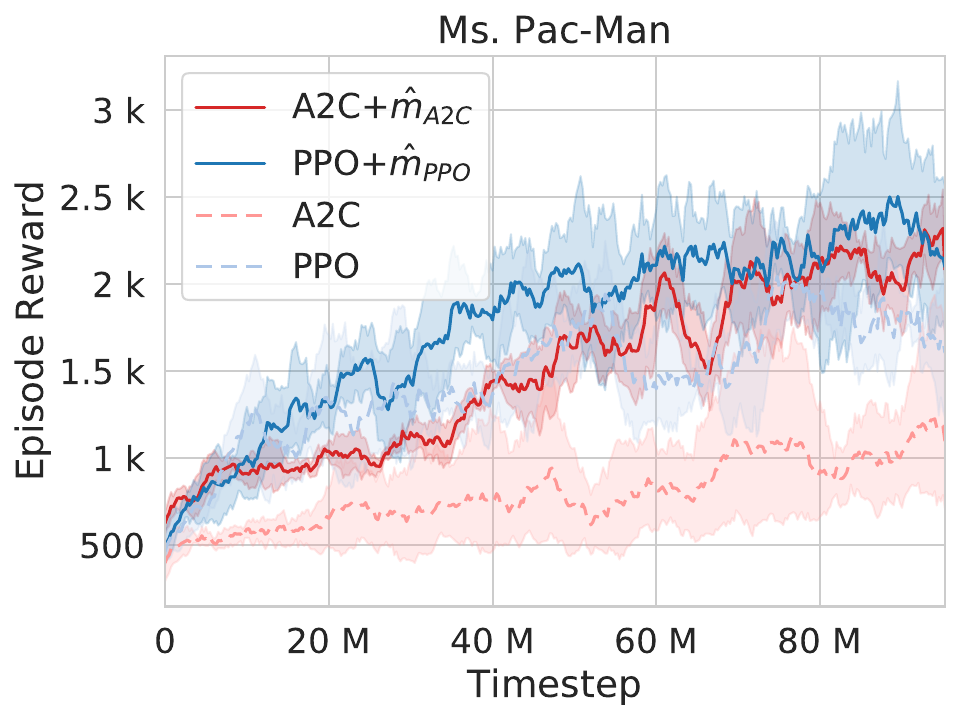}
    \label{fig:learning_curves:compatibility:ms_pacman}
\end{subfigure}
\begin{subfigure}[b]{0.24\linewidth}
    \includegraphics[width=.9\linewidth]{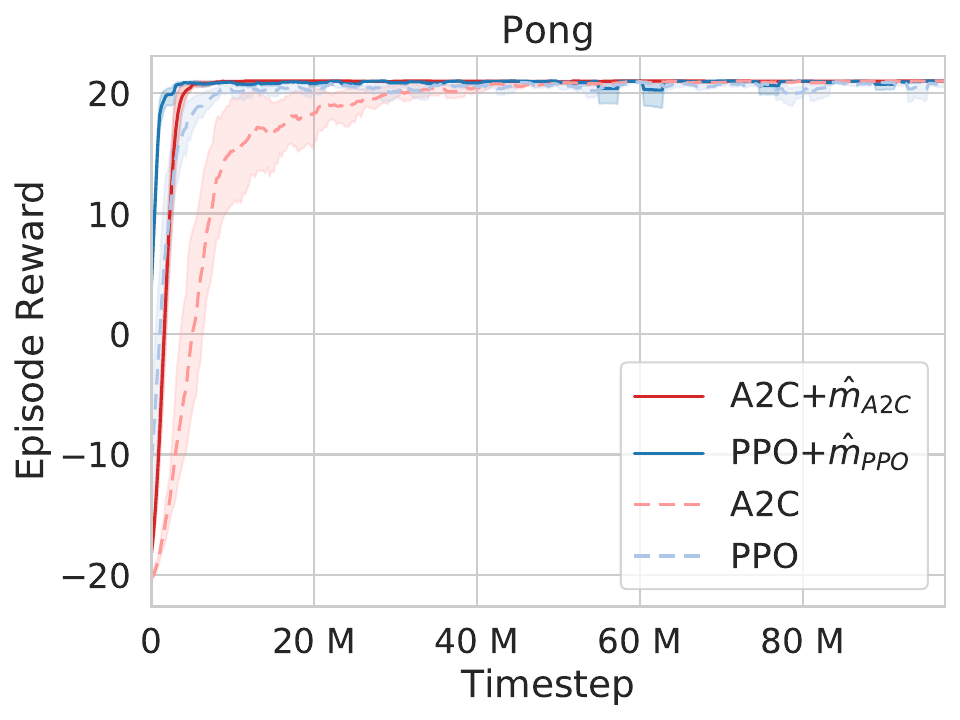}
    \label{fig:learning_curves:compatibility:pong}
\end{subfigure}
\begin{subfigure}[b]{0.24\linewidth}
    \includegraphics[width=.9\linewidth]{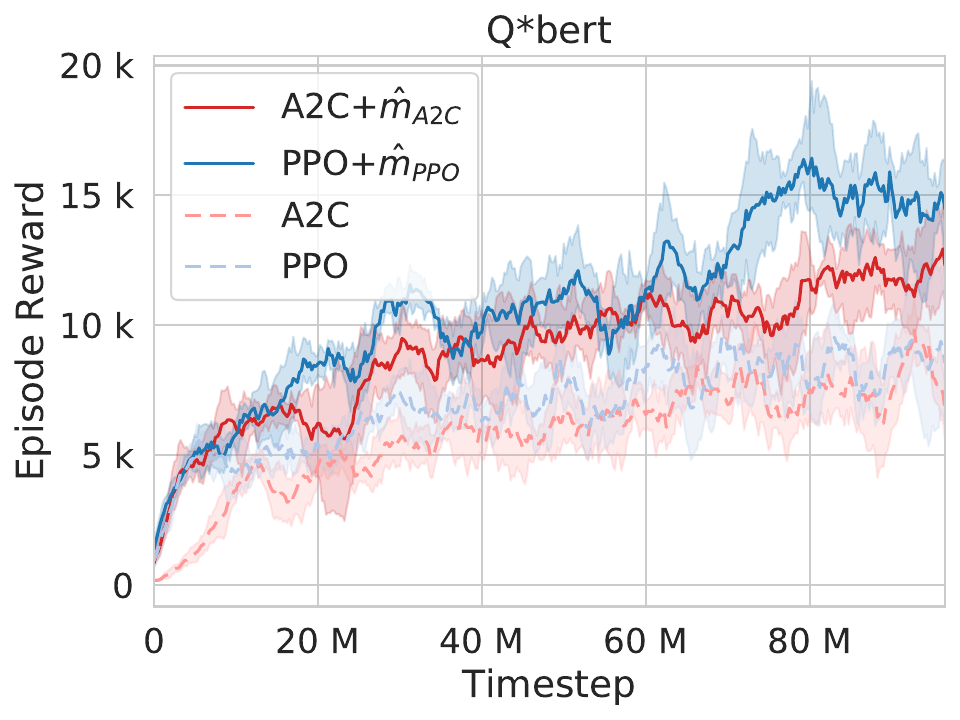}
    \label{fig:learning_curves:compatibility:qbert}
\end{subfigure}
\begin{subfigure}[b]{0.24\linewidth}
    \includegraphics[width=.9\linewidth]{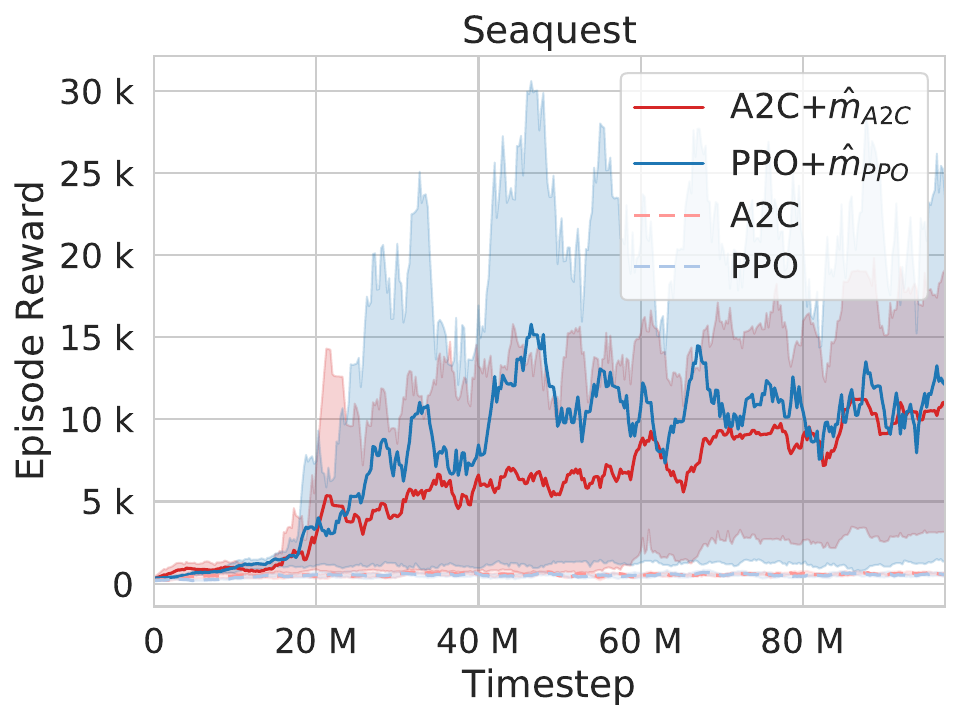}
    \label{fig:learning_curves:compatibility:seaquest}
\end{subfigure}
\caption{
The learning curves of the RL methods with and without the derived macros for validating the proposed methodology. We first perform Algorithm~\ref{alg:genetic_algorithm} with \atwoc{} and \ppo{}, and determine the best macros $\dismacro{\atwoc{}}$ and $\dismacro{\ppo{}}$ for the two RL methods respectively. We then train \atwoc{} using $\augspace{}_{\atwoc{}} = \primspace{} \cup \{ \dismacro{\atwoc{}} \}$ and \ppo{} using $\augspace{}_{\ppo{}} = \primspace{} \cup \{ \dismacro{\ppo{}} \}$ for 100M timesteps.}
\label{fig:learning_curves:compatibility}
\end{figure*}
     \begin{table}[t]
\centering
\caption{Average fitness and improvement (given in parentheses) of each generation using Algorithm~\ref{alg:genetic_algorithm} when training our \atwoc{} agents.}
\label{tab:generation_a2c}
\begin{tabular}{@{}c|rrrr@{}}
\toprule
Generation & \multicolumn{1}{c}{\asteroids{}} & \multicolumn{1}{c}{\beamrider{}} & \multicolumn{1}{c}{\breakout{}} & \multicolumn{1}{c}{\kungfumaster{}} \\ \midrule
0          & 283.42                        & 586.34                        & 17.63                        & 5714.15                          \\
1          & 375.85 (+32.61\%)             & 887.85 (+51.42\%)             & 32.61 (+84.97\%)             & 7662.13 (+34.09\%)               \\
2          & 382.07 (+1.65\%)              & 1007.10 (+13.43\%)            & 54.33 (+66.61\%)             & 8689.08 (+13.40\%)               \\
3          & 504.10 (+31.94\%)             & 1203.91 (+19.54\%)            & 62.36 (+14.78\%)             & 8720.77 (+0.36\%)                \\
4          & 766.49 (+52.05\%)             & 1311.26 (+8.92\%)             & 69.93 (+12.14\%)             & 8720.77 (+0.00\%)                \\
5          & 1024.47 (+33.66\%)            & 1360.44 (+3.75\%)             & 74.89 (+7.09\%)              & 9125.68 (+4.64\%)                \\
6          & 1033.44 (+0.87\%)             & 1404.22 (+3.22\%)             & 74.89 (+0.00\%)              & 9183.96 (+0.64\%)                \\
\bottomrule \toprule
Generation & \multicolumn{1}{c}{\mspacman{}}  & \multicolumn{1}{c}{\pong{}}      & \multicolumn{1}{c}{\qbert{}}    & \multicolumn{1}{c}{\seaquest{}}     \\ \midrule
0          & 431.71                        & -15.91                        & 749.98                       & 203.63                           \\
1          & 451.49 (+4.58\%)              & -14.59 (+8.28\%)              & 1033.72 (+37.83\%)           & 229.49 (+12.70\%)                \\
2          & 481.68 (+6.69\%)              & -13.83 (+5.20\%)              & 1273.18 (+23.16\%)           & 330.41 (+43.97\%)                \\
3          & 504.73 (+4.79\%)              & -13.52 (+2.28\%)              & 1550.30 (+21.77\%)           & 374.55 (+13.36\%)                \\
4          & 537.86 (+6.56\%)              & -13.36 (+1.17\%)              & 1816.05 (+17.14\%)           & 378.33 (+1.01\%)                 \\
5          & 538.24 (+0.07\%)              & -12.72 (+4.74\%)              & 2264.65 (+24.70\%)           & 418.45 (+10.60\%)                \\
6          & 538.24 (+0.00\%)              & -12.72 (+0.00\%)              & 2466.02 (+8.89\%)            & 418.45 (+0.00\%)                 \\
\bottomrule
\end{tabular}
\end{table}

\begin{table}[t]
\centering
\caption{Average fitness and improvement (given in parentheses) of each generation using Algorithm~\ref{alg:genetic_algorithm} when training our \ppo{} agents.}
\label{tab:generation_ppo}
\begin{tabular}{@{}c|rrrr@{}}
\toprule
Generation & \multicolumn{1}{c}{\asteroids{}} & \multicolumn{1}{c}{\beamrider{}} & \multicolumn{1}{c}{\breakout{}} & \multicolumn{1}{c}{\kungfumaster{}} \\ \midrule
0          & 356.47                        & 584.53                        & 38.03                        & 7005.94                          \\
1          & 454.50 (+27.50\%)             & 832.14 (+42.36\%)             & 48.40 (+27.26\%)             & 8597.64 (+22.72\%)               \\
2          & 641.55 (+41.15\%)             & 972.53 (+16.87\%)             & 49.78 (+2.85\%)              & 8994.26 (+4.61\%)                \\
3          & 832.22 (+29.72\%)             & 1020.64 (+4.95\%)             & 52.63 (+5.72\%)              & 9265.07 (+3.01\%)                \\
4          & 1035.22 (+24.39\%)            & 1062.66 (+4.12\%)             & 53.28 (+1.23\%)              & 9386.96 (+1.32\%)                \\
5          & 1243.37 (+20.11\%)            & 1176.72 (+10.73\%)            & 53.28 (+0.00\%)              & 9566.84 (+1.92\%)                \\
6          & 1501.96 (+20.80\%)            & 1176.72 (+0.00\%)             & 53.28 (+0.00\%)              & 9566.84 (+0.00\%)                \\
\bottomrule \toprule
Generation & \multicolumn{1}{c}{\mspacman{}}  & \multicolumn{1}{c}{\pong{}}      & \multicolumn{1}{c}{\qbert{}}    & \multicolumn{1}{c}{\seaquest{}}     \\ \midrule
0          & 534.11                        & 20.27                         & 2277.98                      & 226.30                           \\
1          & 583.46 (+9.24\%)              & 20.41 (+0.71\%)               & 2385.81 (+4.73\%)            & 262.54 (+16.01\%)                \\
2          & 597.63 (+2.43\%)              & 20.62 (+1.04\%)               & 2444.90 (+2.48\%)            & 303.43 (+15.58\%)                \\
3          & 597.72 (+0.02\%)              & 20.71 (+0.41\%)               & 2457.90 (+0.53\%)            & 314.56 (+3.67\%)                 \\
4          & 607.50 (+1.64\%)              & 20.80 (+0.46\%)               & 2464.72 (+0.28\%)            & 408.15 (+29.75\%)                \\
5          & 617.55 (+1.65\%)              & 20.83 (+0.14\%)               & 2482.80 (+0.73\%)            & 418.58 (+2.56\%)                 \\
6          & 617.55 (+0.00\%)              & 20.84 (+0.04\%)               & 2510.49 (+1.12\%)            & 418.58 (+0.00\%)                 \\
\bottomrule
\end{tabular}
\end{table}
 \section{Validation of the Proposed Workflow}
\label{sec:compatibility}

In this section, we validate the proposed workflow presented in Section~\ref{sec:proposed_approach}. We first demonstrate that the macros derived by  Algorithm~\ref{alg:genetic_algorithm} do improve over generations.  We next provide two examples to qualitatively explain the \textit{embedding effect} and the \textit{evaluation effect} of our derived macros in RL. Finally, we quantitatively show that the advantages of the two effects benefit two off-the-shelf RL methods. The \textit{reusability} and \textit{transferability} properties are examined in Section~\ref{sec:transferability_reusability}.

\paragraph{Validation 1: The macros derived over generations.}~\fixtypo{Fig.~\ref{fig:learning_curves:macro_example} uses two different types of environments, \dense{} from \vizdoom{} and \atarienduro{} from \atarifull{},} as example cases to demonstrate the average fitness and improvement of the each generation produced by Algorithm~\ref{alg:genetic_algorithm}.  The primary aim of the table is to offer a verification of Algorithm~\ref{alg:genetic_algorithm} for generating effective macro actions.  The best one is then employed to investigate and examine the \textit{reusability} and \textit{transferability} properties of it.  From the trends of these two example cases, it is observed that the mean episode rewards (i.e., the \textit{fitness}) evaluated by the fitness function (i.e., Algorithm~\ref{alg:evaluation}) improve over generations, revealing that later generations do inherit the advantageous properties from their parents.  Such advantageous properties are retained over generations, pushing the population of the macro actions to evolve toward better fitness. The improving trends therefore suggest that the fitness function presented in Algorithm~\ref{alg:evaluation} is effective and reliable for Algorithm~\ref{alg:genetic_algorithm}. For the other environments and RL methods, the average fitness of the population for each generation is provided in Tables~\ref{tab:generation_a2c}~and~\ref{tab:generation_ppo}.

\paragraph{Validation 2: The embedding effect of the derived macros.}~In our second validation, we employ ``\dense{}'' from \vizdoom{} to explore and discuss the benefits of our derived macros in terms of the \textit{embedding effect} by comparing the macros derived by Algorithm~\ref{alg:genetic_algorithm} with the action repeat macro, in which the same primitive actions are repeatedly executed.  To this end, we first use the proposed macro generation stage to construct the best macro $\dismacro{\denseabbr{}}$ for ``\dense{}''. Then, in order to construct a proper action repeat macro $\dismacro{\methodrepeat{}}$, we evaluate all possible action repeat macros with the same length equal to $\dismacro{\denseabbr{}}$ for 10M timesteps. The macro with the highest evaluation score is then selected as $\dismacro{\methodrepeat{}}$. The final constructed $\dismacro{\denseabbr{}}$ and $\dismacro{\methodrepeat{}}$ are (\texttt{MOVE\_FORWARD}, \texttt{MOVE\_FORWARD}, \texttt{TURN\_RIGHT}) and (\texttt{MOVE\_FORWARD}, \texttt{MOVE\_FORWARD}, \texttt{MOVE\_FORWARD}), respectively. To visualize the impacts of them, we plot and compare the learning curve of ``\curiosity{}+$\dismacro{\denseabbr{}}$'' against that of ``\curiosity{}+$\dismacro{\methodrepeat{}}$'' in Fig.~\ref{fig:learning_curves:macro_example:dense}.  Fig.~\ref{fig:learning_curves:macro_example:dense} further includes a curve for the vanilla ``\curiosity{}'' for the purpose of comparison.  
It is observed that the curve of ``\curiosity{}+$\dismacro{\methodrepeat{}}$'' is worse than the curves of ``\curiosity{}'' and ``\curiosity{}+$\dismacro{\denseabbr{}}$'' in the early stage of the training phase.  This observation provides two insights.  First, although both $\dismacro{\denseabbr{}}$ and $\dismacro{\methodrepeat{}}$ allow the RL agents to bypass intermediate states by performing consecutive actions, $\dismacro{\methodrepeat{}}$ does not lead to immediate positive impacts when compared with the vanilla ``\curiosity{}''.  This suggests that not all macro generation methods are able to construct a macro action that benefits equally from the \textit{embedding effect}.  Second, it is observed that $\dismacro{\denseabbr{}}$ enables the agent to perform better than the vanilla ``\curiosity{}'', indicating that the macro action derived by our macro generation stage does provide positive impact when the agent is allowed to bypass intermediate states during the training phase of it.

\paragraph{Validation 3: The evaluation effect of the derived macros.}~In addition to the \textit{embedding effect} discussed in the former example, in the third validation, we employ the ``\atarienduro{}'' environment from \atarifull{} to explain why the derived macro can benefit from the \textit{evaluation effect}. In ``\atarienduro{}'', the agent controls a car to race with the other rival cars, and receives a reward signal only when it passes any one of them. We illustrate the learning curves in Fig.~\ref{fig:learning_curves:macro_example:enduro}, and use them to compare the \atwoc{} agent trained with and without the best macro $\dismacro{\atwoc{}}$ constructed by our methodology for 100M timesteps. According to the results, $\dismacro{\atwoc{}}$ is $(\texttt{FIRE}, \texttt{FIRE})$, corresponding to two repeated forward moves. This macro enables the agent to learn to surpass the rival cars in the environment easier. It is observed that \atwoc{} with $\dismacro{\atwoc{}}$ outperforms the vanilla \atwoc{}, which is hardly able to learn an effective policy throughout the training process.  Cluelessly performing two consecutive forward moves poses a risk for the car to hit the rival cars, thus preventing the vanilla \atwoc{} from learning an effective policy.  This observation reveals that the derived macro improves the search guidance of the RL agent due to the altered value estimation of each state.  This is a direct advantage offered by the \textit{embedding effect}. 

\paragraph{Validation 4: Examination of the derived macro actions.}~The macro action generated using our methodology, which has the highest fitness score within the population in a generation, is defined as the best macro. In order to examine whether the best macros $\hat{m}$ derived by our methodology is able to benefit the selected RL methods, we first perform Algorithm~\ref{alg:genetic_algorithm} with \atwoc{} and \ppo{}, and determine the best macros $\dismacro{\atwoc{}}$ and $\dismacro{\ppo{}}$ for the two RL methods respectively. We then train \atwoc{} using $\augspace{}_{\atwoc{}} = \primspace{} \cup \{ \dismacro{\atwoc{}} \}$ and \ppo{} using $\augspace{}_{\ppo{}} = \primspace{} \cup \{ \dismacro{\ppo{}} \}$ for 100M timesteps. The training timesteps are chosen to be much longer than the ``fitness phase’’ of Algorithm~\ref{alg:genetic_algorithm} to highlight the impact of the best macros on the agents in the long run.  The results shown in Fig.~\ref{fig:learning_curves:compatibility} indicate that \revision{the RL methods benefit from the macros derived by the proposed methodology.} We list the best macros for the \atarifull{} environments considered in this paper in Table~\ref{tab:atari_constructed_macros} as a reference. These best macros are then used in Section~\ref{sec:transferability_reusability} for validating the \textit{reusability} and the \textit{transferability}  properties.
     \begin{table}[t]
\centering
\caption{List of the macro actions generated for each \atarifull{} environment using our methodology.}
\label{tab:atari_constructed_macros}
\begin{tabular}{c|cc}
\toprule
Environment         & $\dismacro{\atwoc{}}$ \\
\midrule
\asteroids{}        & (\texttt{RIGHT}, \texttt{FIRE}, \texttt{LEFT}, \texttt{NOOP}) \\
\beamrider{}        & (\texttt{FIRE}, \texttt{NOOP}, \texttt{FIRE}, \texttt{FIRE}) \\
\breakout{}         & (\texttt{LEFT}, \texttt{LEFT}, \texttt{NOOP}, \texttt{RIGHT}) \\
\kungfumaster{}     & (\texttt{DOWNLEFTFIRE}, \texttt{RIGHTFIRE}, \texttt{DOWNLEFTFIRE}, \texttt{DOWNLEFT}) \\
\mspacman{}         & (\texttt{RIGHT}, \texttt{NOOP}, \texttt{RIGHT}, \texttt{NOOP}, \texttt{RIGHT}, \texttt{NOOP}) \\
\pong{}             & (\texttt{LEFT}, \texttt{NOOP}, \texttt{LEFT}) \\
\qbert{}            & (\texttt{NOOP}, \texttt{RIGHT}, \texttt{DOWN}, \texttt{DOWN}) \\
\seaquest{}         & (\texttt{UP}, \texttt{RIGHT}, \texttt{LEFT}, \texttt{NOOP}) \\
\bottomrule
\toprule
Environment         & $\dismacro{\ppo{}}$ \\
\midrule
\asteroids{}        & (\texttt{NOOP}, \texttt{FIRE}, \texttt{FIRE}, \texttt{NOOP}, \texttt{FIRE}) \\
\beamrider{}        & (\texttt{FIRE}, \texttt{FIRE}, \texttt{RIGHT}, \texttt{LEFT}) \\
\breakout{}         & (\texttt{LEFT} , \texttt{NOOP}, \texttt{LEFT}, \texttt{RIGHT}) \\
\kungfumaster{}     & (\texttt{UPLEFTFIRE}, \texttt{DOWN}, \texttt{DOWNLEFTFIRE}) \\
\mspacman{}         & (\texttt{NOOP}, \texttt{DOWN}) \\
\pong{}             & (\texttt{LEFT}, \texttt{RIGHT}, \texttt{LEFT}) \\
\qbert{}            & (\texttt{DOWN}, \texttt{DOWN}, \texttt{RIGHT}, \texttt{DOWN}) \\
\seaquest{}         & (\texttt{UP}, \texttt{FIRE}, \texttt{DOWN}, \texttt{UP}) \\
\bottomrule
\end{tabular}
\end{table}
     \begin{figure*}[t]
\centering
\begin{subfigure}[b]{0.24\linewidth}
    \includegraphics[width=.9\linewidth]{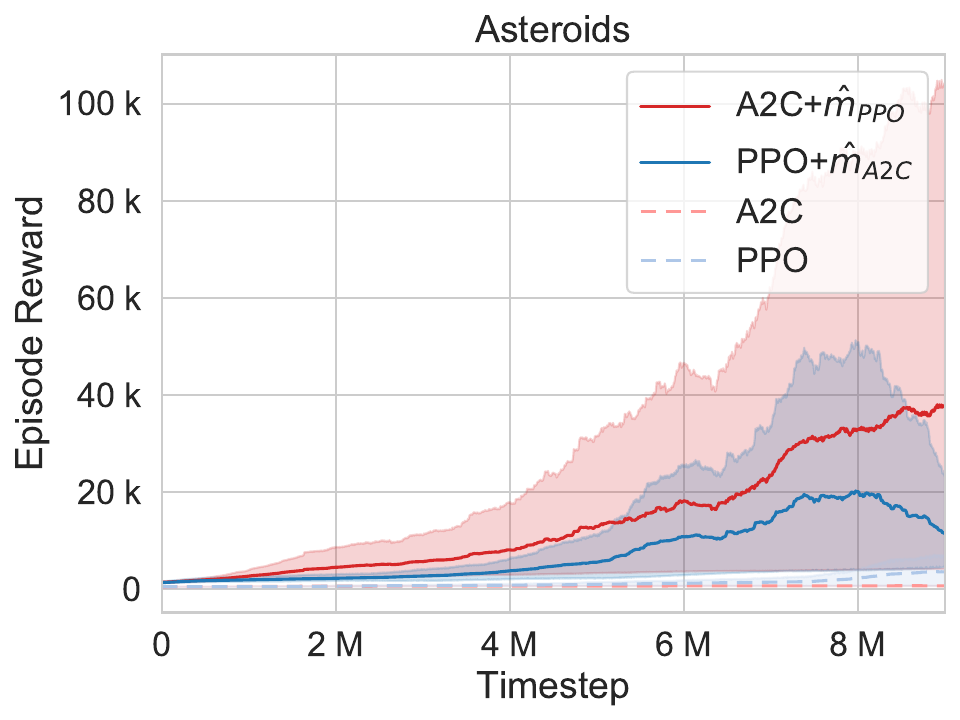}
    \label{fig:learning_curves:transferability:asteroids}
\end{subfigure}
\begin{subfigure}[b]{0.24\linewidth}
    \includegraphics[width=.9\linewidth]{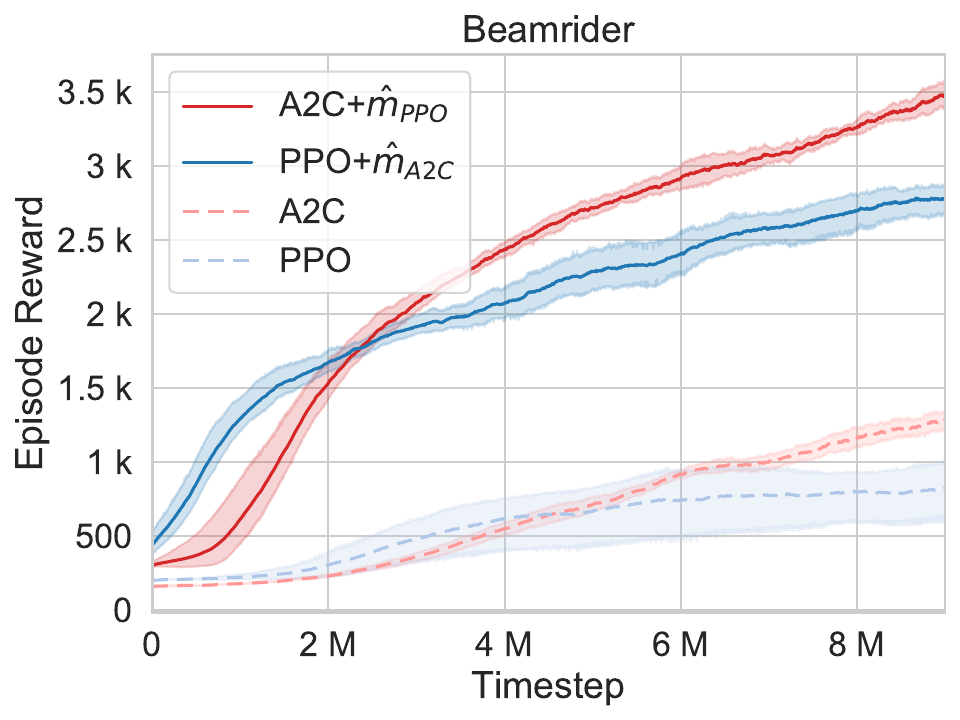}
    \label{fig:learning_curves:transferability:beam_rider}
\end{subfigure}
\begin{subfigure}[b]{0.24\linewidth}
    \includegraphics[width=.9\linewidth]{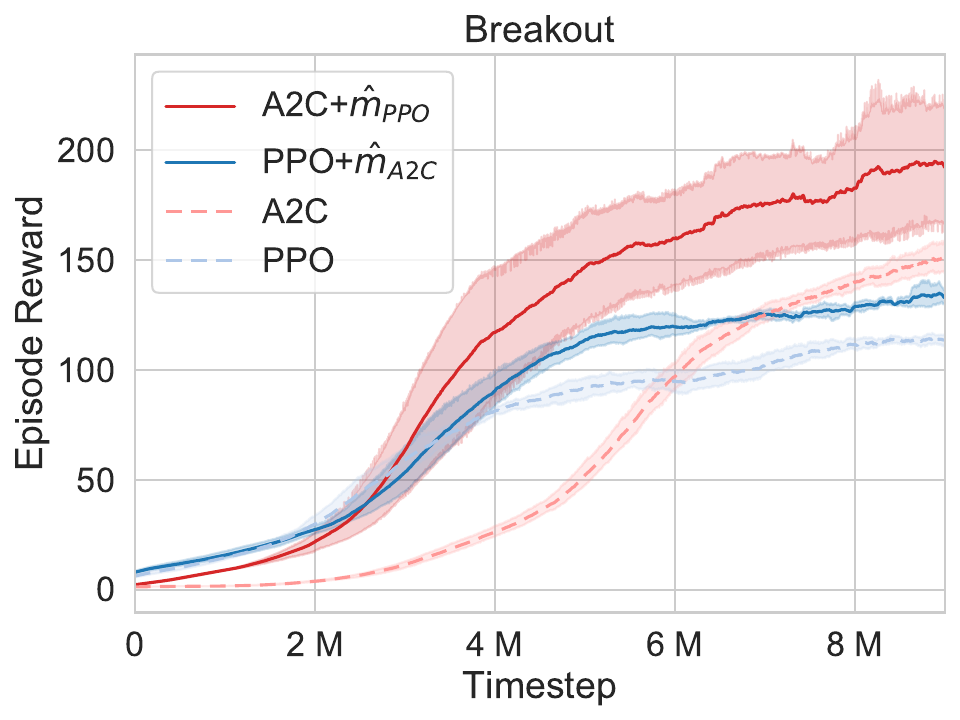}
    \label{fig:learning_curves:transferability:breakout}
\end{subfigure}
\begin{subfigure}[b]{0.24\linewidth}
    \includegraphics[width=.9\linewidth]{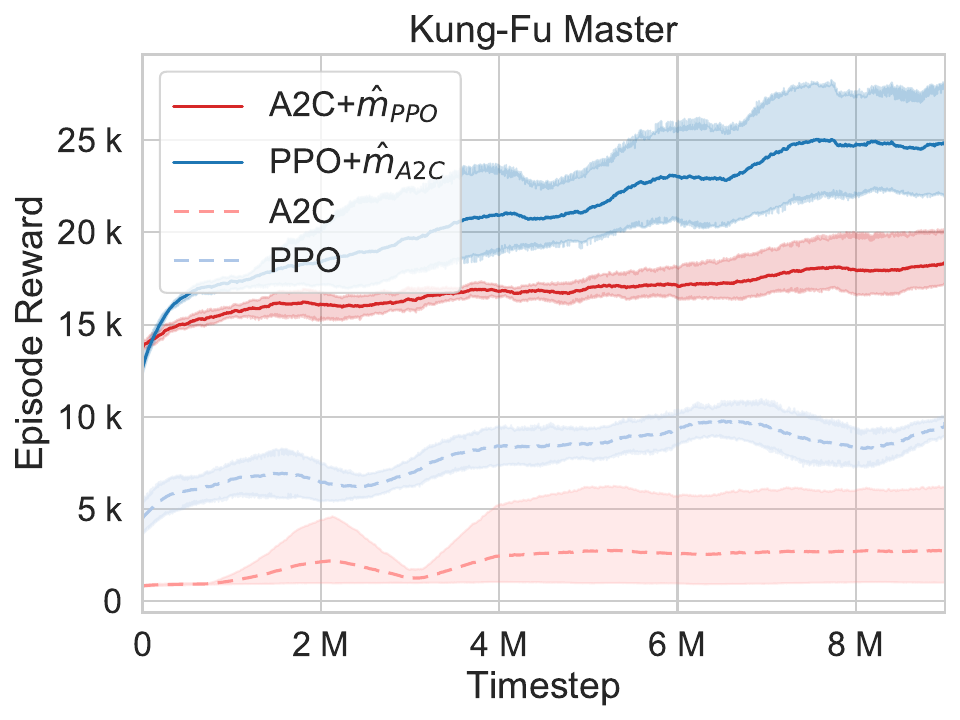}
    \label{fig:learning_curves:transferability:kung_fu_master}
\end{subfigure}
\begin{subfigure}[b]{0.24\linewidth}
    \includegraphics[width=.9\linewidth]{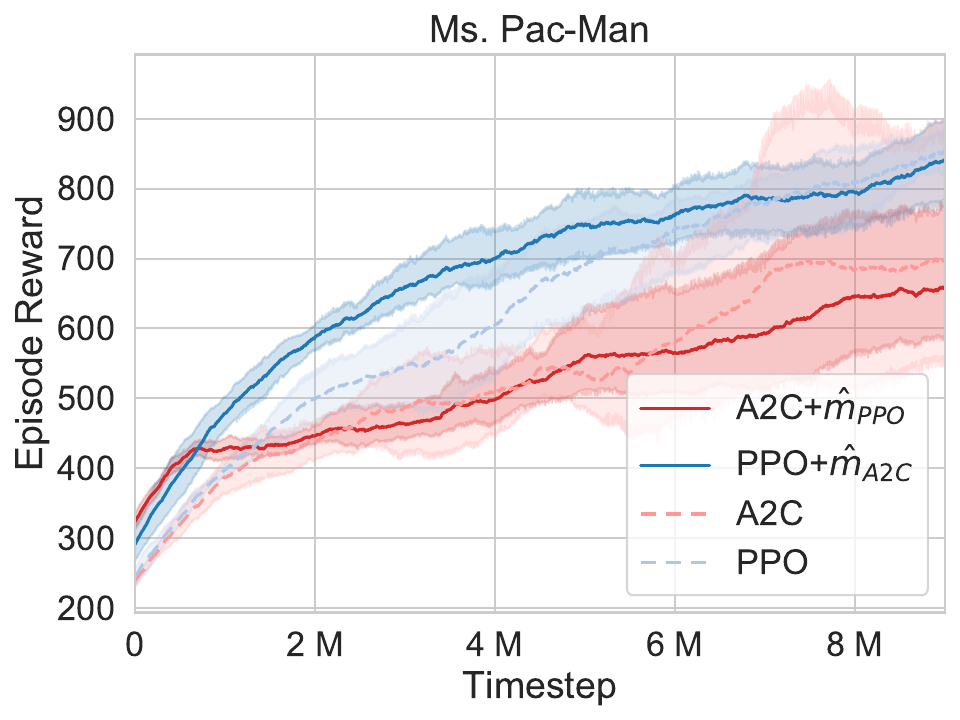}
    \label{fig:learning_curves:transferability:ms_pacman}
\end{subfigure}
\begin{subfigure}[b]{0.24\linewidth}
    \includegraphics[width=.9\linewidth]{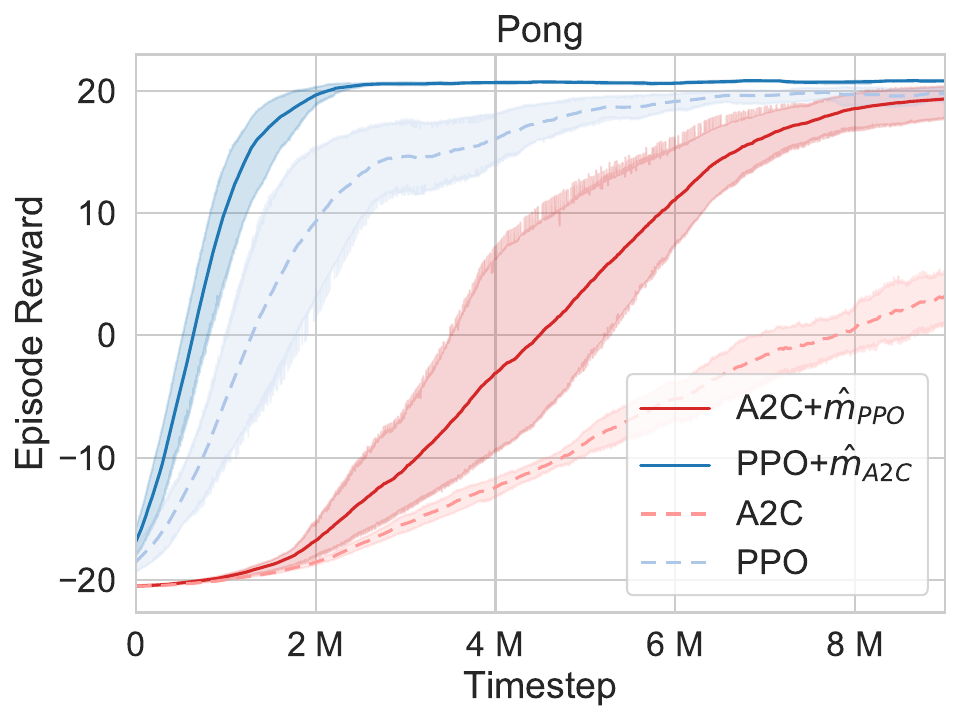}
    \label{fig:learning_curves:transferability:pong}
\end{subfigure}
\begin{subfigure}[b]{0.24\linewidth}
    \includegraphics[width=.9\linewidth]{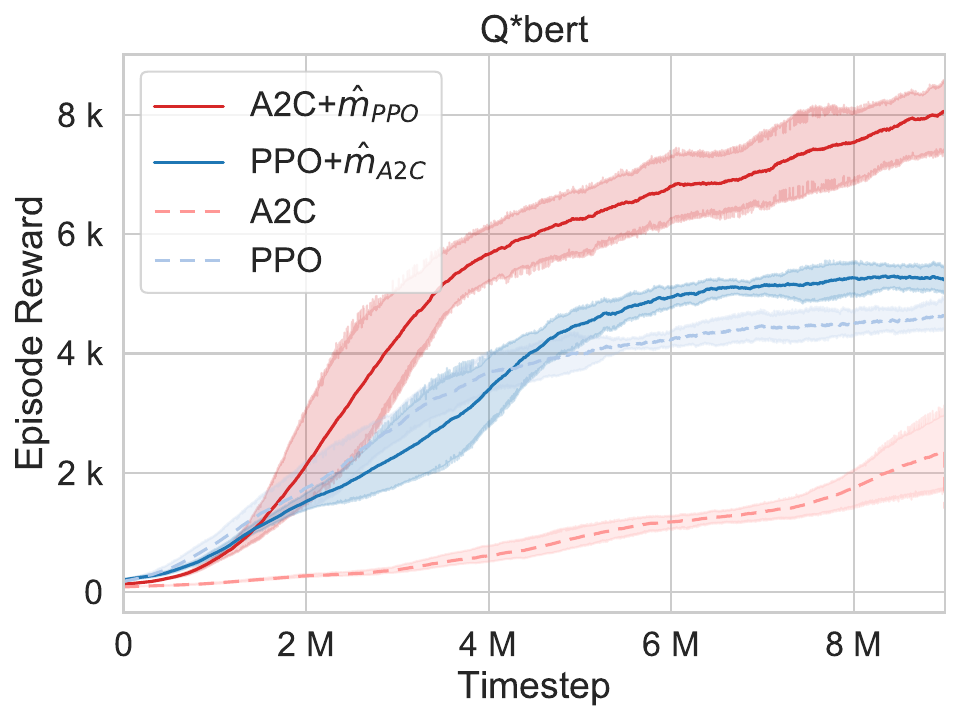}
    \label{fig:learning_curves:transferability:qbert}
\end{subfigure}
\begin{subfigure}[b]{0.24\linewidth}
    \includegraphics[width=.9\linewidth]{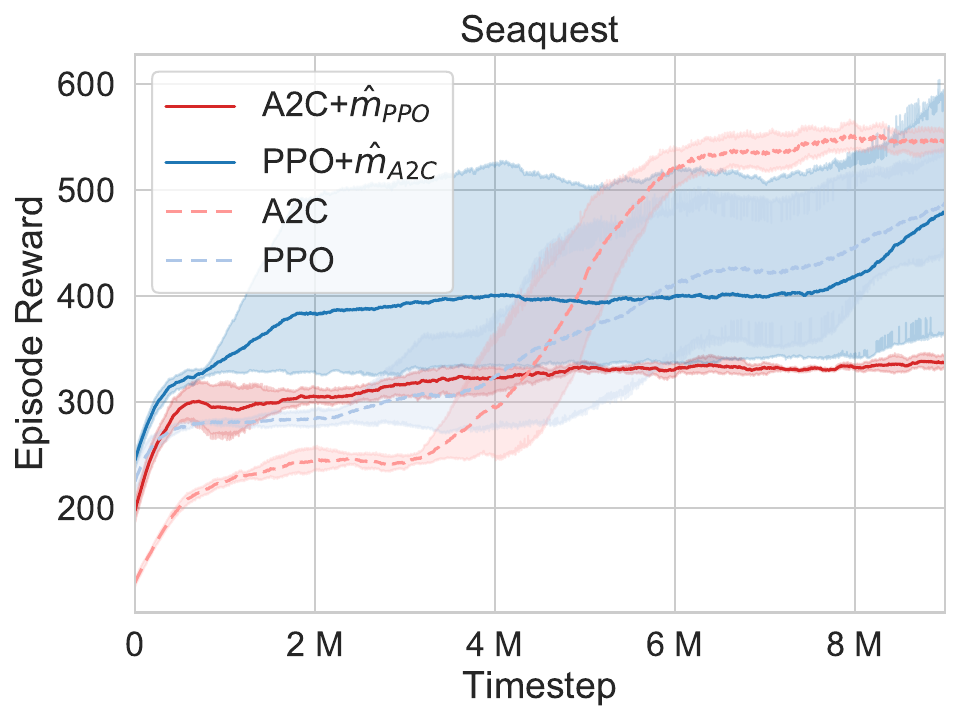}
    \label{fig:learning_curves:transferability:seaquest}
\end{subfigure}
\caption{
The learning curves of the RL methods with and without the derived macros for evaluating \textit{reusability}. We reuse the macros constructed by Algorithm~\ref{alg:genetic_algorithm} to train an \atwoc{} agent and a \ppo{} agent using $\augspace{}_{\ppo{}}$ and $\augspace{}_{\atwoc{}}$, respectively, for 10M timesteps.
}
\label{fig:learning_curves:transferability}
\end{figure*}
     \begin{figure*}[t]
\centering
\begin{subfigure}[b]{0.24\linewidth}
    \includegraphics[width=.9\linewidth]{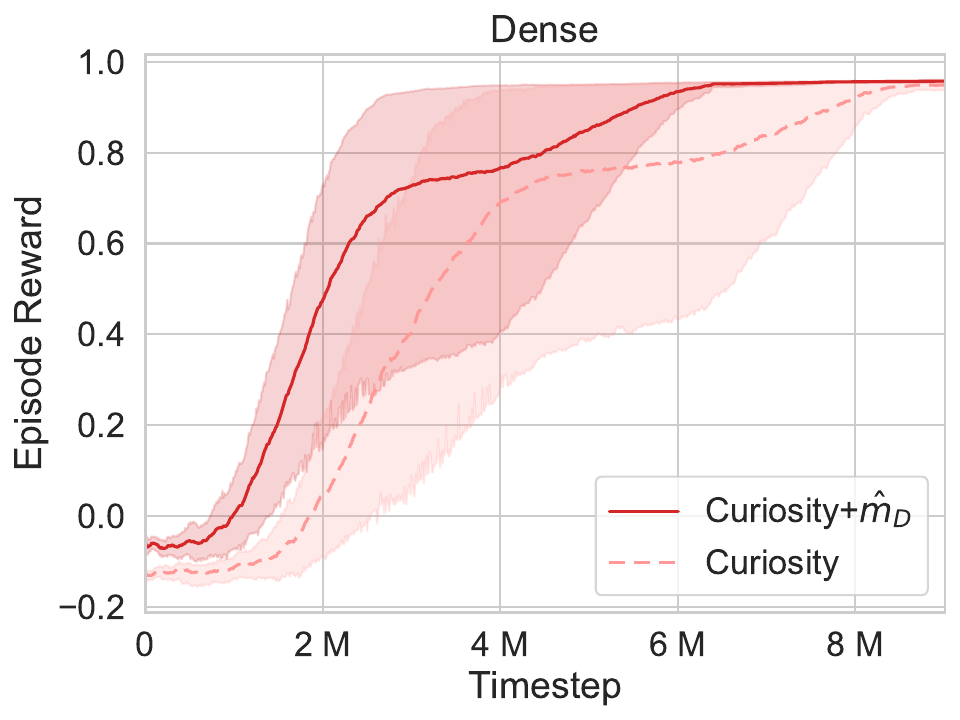}
    \label{fig:learning_curves:reusability:dense}
\end{subfigure}
\begin{subfigure}[b]{0.24\linewidth}
    \includegraphics[width=.9\linewidth]{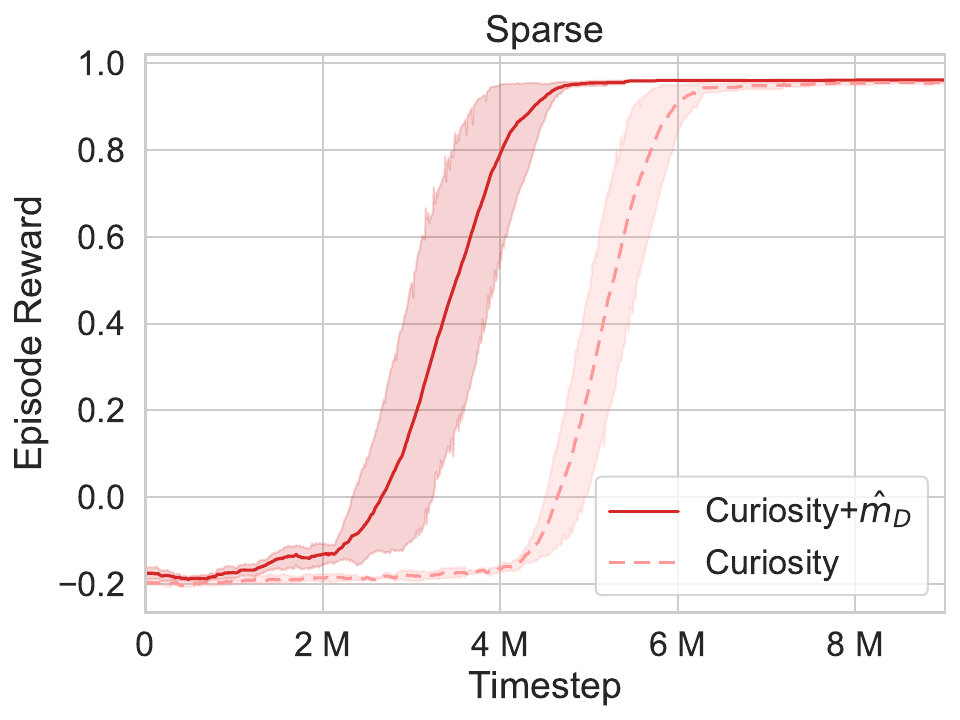}
    \label{fig:learning_curves:reusability:sparse}
\end{subfigure}
\begin{subfigure}[b]{0.24\linewidth}
    \includegraphics[width=.9\linewidth]{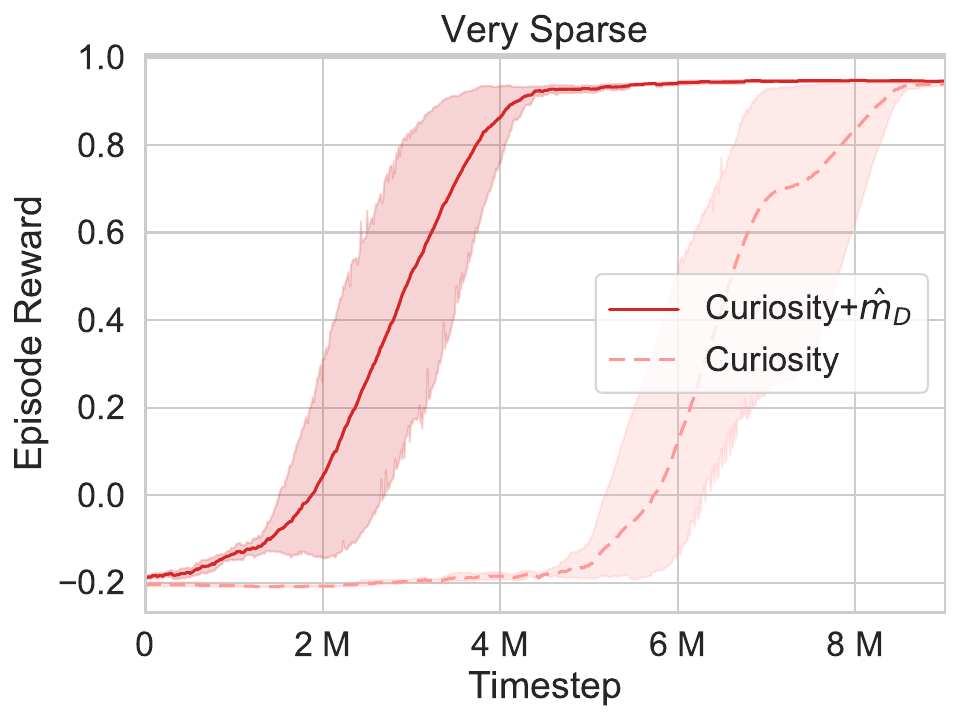}
    \label{fig:learning_curves:reusability:very_sparse}
\end{subfigure}
\begin{subfigure}[b]{0.24\linewidth}
    \includegraphics[width=.9\linewidth]{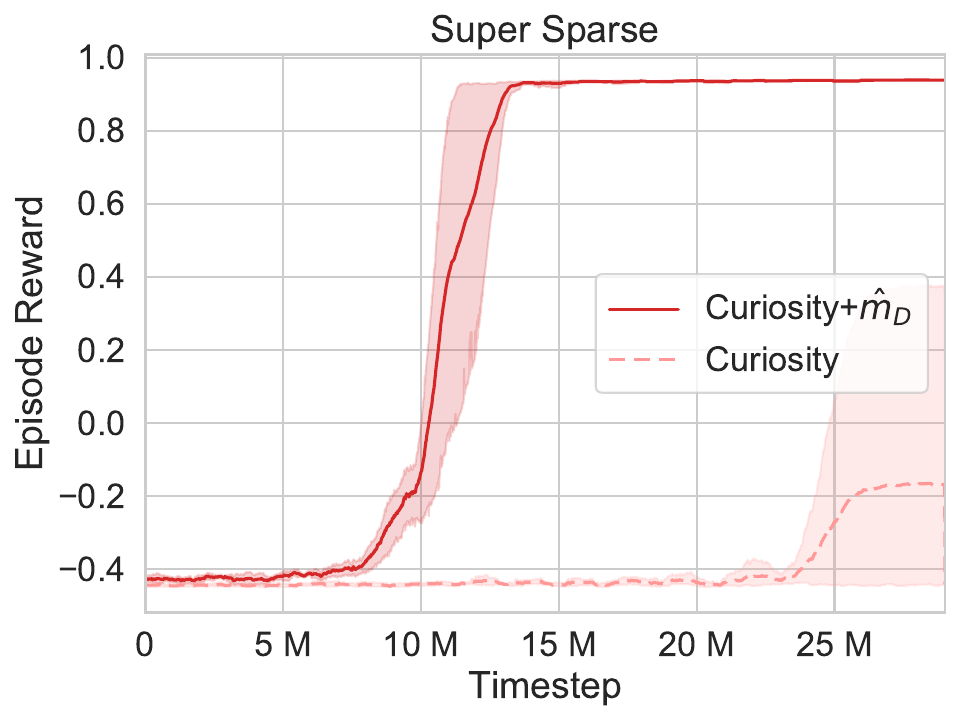}
    \label{fig:learning_curves:reusability:super_sparse}
\end{subfigure}
\caption{
The learning curves of the RL method with and without the derived macro for evaluating the \textit{transferability} property. We utilize the constructed macro presented in Section~\ref{sec:compatibility} to validate the \textit{transferability} of it in the \sparse{}, \verysparse{}, and \supersparse{} \vizdoom{} tasks. For the \supersparse{} task, `\curiosity{}+$\dismacro{\denseabbr{}}$' converges at around 13M timesteps, while `\curiosity{}' just begins to learn at about 25M timesteps.
}
\label{fig:learning_curves:reusability}
\end{figure*}
     \begin{table}[t]
\centering
\caption{The evaluation results of the four \vizdoom{} tasks. We present the mean timesteps required by the agents to reach the goal over 100 episodes. The numerical results show that the agents trained with \curiosity{}+$\dismacro{\denseabbr{}}$ require less time than those trained with \curiosity{} for all the four \vizdoom{} tasks.}
\label{tab:doom_evaluation_results}
\begin{tabular}{l|rrrr}
\toprule
Task            & \curiosity{} & \curiosity{}+$\dismacro{\denseabbr{}}$   & \textbf{Reduction (\%)} \\
\midrule
\dense{}        & 120.43    & 94.52                                 & \textbf{-21.51}  \\
\sparse{}       & 157.46    & 139.69                                & \textbf{-11.29}  \\
\verysparse{}   & 237.45    & 161.82                                & \textbf{-31.85}  \\
\supersparse{}  & 405.63    & 223.95                                & \textbf{-44.79}  \\
\bottomrule
\end{tabular}
\end{table}
 \section{Reusability and Transferability  Properties}
\label{sec:transferability_reusability}

In this section, we first present a set of experiments to validate the fact that the macros derived by our workflow might exhibit the \textit{reusability} property. We then discuss the  existence of the \textit{transferability} property for the macro presented in Section~\ref{sec:compatibility} in similar environments with different reward settings.

\paragraph{The Reusability Property.}~The \textit{reusability} property is said to exist if a macro action constructed along with one RL method can be used by another RL method for training. To validate the existence of this property, we reuse the macros listed in Table~\ref{tab:atari_constructed_macros}  to train an \atwoc{} agent and a \ppo{} agent using $\augspace{}_{\ppo{}}$ and $\augspace{}_{\atwoc{}}$, respectively, for 10M timesteps. We plot the results of \asteroids{}, \beamrider{}, \breakout{}, \kungfumaster{}, \mspacman, \pong{}, \qbert{}, and \seaquest{} in Fig.~\ref{fig:learning_curves:transferability}. The experimental results show that the \atwoc{} and \ppo{} agents are able to be benefited from the provided macros in most cases. This justifies the existence of the \textit{reusability} property of the macros. The above observations also suggest that the macro actions derived by our workflow could exhibit invariance when they are employed by different RL methods during the training phase of the agents.  This property further implies that once a good macro action is derived by the workflow, it can be later leveraged in the training phase of another RL method and reduce the time required by the macro generation phase.

\paragraph{The Transferability Property.}~The \textit{transferability} property is said to exist if the constructed macros can be leveraged in similar environments with different reward settings. In order to confirm this property, we utilize the macro $\dismacro{\denseabbr{}}=$ (\texttt{MOVE\_FORWARD}, \texttt{MOVE\_FORWARD}, \texttt{TURN\_RIGHT}) derived from the \dense{} reward setting presented in Section~\ref{sec:compatibility} to validate the \textit{transferability} of it in tasks with sparse reward settings, including \sparse{}, \verysparse{}, and \supersparse{}. The results are plotted in Fig.~\ref{fig:learning_curves:reusability}. We also provide the numerical results in Table~\ref{tab:doom_evaluation_results}. Fig.~\ref{fig:learning_curves:reusability} demonstrates that the agents with $\dismacro{\denseabbr{}}$ learn relatively faster than the agents without it. Fig.~\ref{fig:learning_curves:reusability} also reveals that the gap between ``\curiosity{}+$\dismacro{\denseabbr{}}$'' and ``\curiosity{}'' grows as the sparsity of the reward signal increases. For the \supersparse{} task, ``\curiosity{}+$\dismacro{\denseabbr{}}$'' converges at around 13M timesteps, while ``\curiosity{}'' just begins to learn at about 25M timesteps. These results thus validate the \textit{transferability} property of the derived macro action, and suggests that the macro derived in an environment can be utilized in a similar one with different reward settings, even if the reward signal becomes sparser than the original one.  This is also due to the benefits offered by the \textit{embedding} and the \textit{evaluation effects}. The former reduces the search depth, while the latter provides better search guidance. These two effects together enable the RL agent to quickly reach the goal and updates its value estimates of the states more efficiently than the vanilla case.
 \section{Conclusions}
\label{sec:conclusions}

In this paper, we have presented a methodology to examine the \textit{reusability} and the \textit{transferability} properties of the derived macros in the RL domain.  We presented a workflow to generate macros, and utilized them with various evaluation configurations.We first validated the workflow, and showed that the derived macros exhibit the \textit{embedding} and \textit{evaluation effects}.  We then examined the \textit{reusability} property between RL methods and the \textit{transferability} property among similar environments. Since the optimization target of the fitness phase is to find macros that can benefit the performance of the RL agents, the \textit{reusability} and \textit{transferability} properties found in most of the experiments can thus serve as the evidence that they are indeed possessed by the macros discovered by the proposed algorithm.
The macros possessing the properties can potentially save the effort of the macro generation procedure.
 \section*{Acknowledgement}

This work was supported by the Ministry of Science and Technology (MOST) in Taiwan under grant number MOST 111-2628-E-007-010. The authors acknowledge the financial support from MediaTek Inc., Taiwan. The authors would also like to acknowledge the donation of the GPUs from NVIDIA Corporation and NVIDIA AI Technology Center (NVAITC) used in this research work.

\bibliographystyle{ACM-Reference-Format}
\bibliography{references}

\end{document}